\documentclass[10pt,twocolumn,letterpaper]{article}

\usepackage[pagenumbers]{cvpr}

\usepackage[most]{tcolorbox}
\usepackage{booktabs}
\usepackage{colortbl}
\usepackage{algorithm}
\usepackage{algpseudocode}

\usetikzlibrary{patterns,decorations.pathmorphing}
\definecolor{promptband}{HTML}{4A7FC1}
\definecolor{phblue}{HTML}{1F5FA8}
\definecolor{phteal}{HTML}{12908E}
\definecolor{phpurple}{HTML}{5B2C9E}
\newcommand{\ph}[2]{\textcolor{#1}{\textbf{#2}}}
\tikzset{sketch/.style={decorate,decoration={random steps,segment length=9pt,amplitude=0.45pt}}}

\definecolor{cvprblue}{rgb}{0.21,0.49,0.74}
\definecolor{projectpink}{RGB}{219,80,140}
\usepackage[pagebackref,breaklinks,colorlinks,allcolors=cvprblue]{hyperref}
\makeatletter
\renewcommand{\theHALG@line}{\thealgorithm.\arabic{ALG@line}}
\makeatother

\def\confName{CVPR}
\def\confYear{2027}

\title{MirrorWorld: Taming Video Diffusion Models for Mirror Reflection Generation}

\author{
Youjun Zhao\textsuperscript{1}\quad
Alex Warren\textsuperscript{2}\quad
Gary K.L. Tam\textsuperscript{2}\quad
Rynson W.H. Lau\textsuperscript{1}\\
\textsuperscript{1}City University of Hong Kong\quad
\textsuperscript{2}Swansea University\\
{\tt\small\href{https://youjunzhao.github.io/MirrorWorld/}{\textcolor{projectpink}{https://youjunzhao.github.io/MirrorWorld/}}}
}

\begin{document}

\twocolumn[{%
\renewcommand\twocolumn[1][]{#1}%
\vspace*{-26pt}%
\maketitle
\vspace{-3em}%
\begin{center}
    \centering
    \captionsetup{type=figure}
    \includegraphics[width=\textwidth]{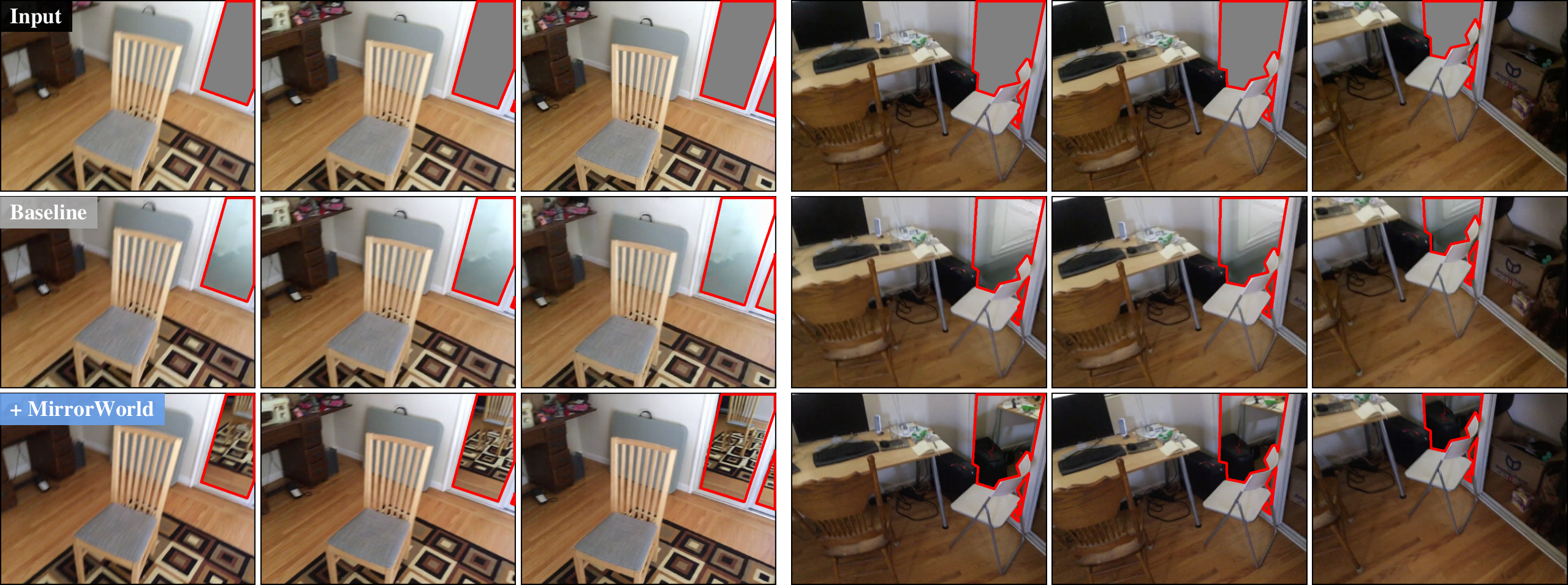}
    \captionof{figure}{Video mirror reflection generation remains challenging for VDMs. While the baseline VDM fails to reflect the visible scene, MirrorWorld synthesizes accurate objects and spatial layouts inside masked mirror regions. Red lines outline the input mirror mask.}
    \label{teasor}
\end{center}
\vspace{1em}%
}]

\begin{abstract}
Recent advances in video diffusion models (VDMs) have enabled high-fidelity video synthesis. However, generating mirror reflections remains challenging because the content within a mirror must remain consistent with the surrounding scene. Existing VDMs are not specifically designed to model scene-to-mirror relationships, and may produce reflections with incorrect content or inconsistent spatial arrangements. To address this, we introduce video mirror reflection generation, a novel task of synthesizing plausible reflections within masked mirror regions. We observe that mirror reflection generation involves two complementary challenges: determining what scene content should be reflected and how the reflected content should be spatially arranged within the mirror region. Motivated by this observation, we propose \textbf{MirrorWorld}, a reflection-aware video inpainting framework that models scene-to-mirror relationships during generation. Specifically, we introduce \textbf{Semantic Relation Distillation (SRD)}, which transfers relational priors from a frozen visual foundation model to encourage semantic associations between visible scene content and mirror regions. We further propose \textbf{Geometric Transformation Alignment (GTA)}, which learns a transformation that guides the spatial arrangement of reflected content. The two components play complementary roles, with SRD modeling what should be reflected and GTA modeling how it should be arranged. To facilitate research on this problem, we construct the first benchmark for video mirror reflection generation by repurposing four existing video mirror datasets into a unified reflection reconstruction task. Experimental results show that MirrorWorld achieves superior reflection reconstruction quality over representative image-based reflection generation methods and strong video inpainting baselines.
\end{abstract}

\section{Introduction}
\label{sec:intro}

Recent advances in video diffusion models (VDMs)~\cite{cogvideox,wan,hunyuanvideo,LTXVideo} have enabled high-fidelity video synthesis with realistic objects, complex motion, and smooth camera movement. These capabilities support applications such as movie creation~\cite{captaincinema}, synthetic data generation~\cite{panacea}, and interactive environment simulation~\cite{genie}. At the same time, generating consistent content across different regions of a scene remains a significant challenge. Mirror reflections serve as a representative case, as the content in a mirror is constrained by its relationship with the surrounding scene rather than being determined solely from local visual context. This raises a natural question: can VDMs fill a mirror with reflections that are consistent with the visible scene?

To investigate this question, we apply VACE~\cite{vace}, a representative VDM, to fill masked mirror regions in real videos with visible surrounding scene. As shown in Figure~\ref{teasor}, VACE fails to reflect the visible scene, instead filling the mirror regions with unrelated surfaces and omitting the foreground objects. These observations indicate that mirror reflection generation remains challenging for VDMs, even when the surrounding scene is explicitly given.

A recent study has explored mirror reflection generation in image diffusion models~\cite{mirrorfusion} by formulating the task as an image inpainting problem. However, it targets synthetic, object-centric scenarios where a single isolated foreground object is placed in front of a mirror. This simplified setting fails to capture real-world scenes, where mirrors reflect environments containing multiple objects and complex background structures. Furthermore, applying an image model frame-by-frame to videos fails to maintain consistent reflections under camera motion and scene dynamics. Extending reflection generation to real-world videos therefore requires accurately inferring complex scene-to-mirror relationships while maintaining coherent reflections across frames.

In this work, we study mirror reflection generation in real-world videos by reconstructing mirror reflections within masked mirror regions. This formulation isolates reflection reasoning from unconstrained video generation while providing reliable targets for evaluation.
Mirror reflection generation differs from conventional video inpainting in two ways. First, mirror content is not arbitrary in our setting and should be inferred from reflection-relevant scene content. Second, reflected content should be arranged consistently within the mirror region. This requires addressing two complementary challenges: \emph{what} should be reflected, and \emph{how} it should be arranged.

To address these challenges, we propose \textbf{MirrorWorld}, a reflection-aware video inpainting framework that models scene-to-mirror relationships.
Given a video with masked mirror regions, MirrorWorld reconstructs missing reflections while preserving non-mirror content by aligning scene features and mirror-region features.
MirrorWorld consists of two components. \textbf{Semantic Relation Distillation (SRD)} transfers relational information from a frozen visual foundation model to encourage mirror-region representations to associate with reflection-relevant content in the visible scene. \textbf{Geometric Transformation Alignment (GTA)} learns a transformation that guides the spatial arrangement of reflected content. Intuitively, SRD addresses \emph{what} should be reflected, while GTA addresses \emph{how} it should be arranged, together encouraging scene-consistent reflection generation.

To facilitate future research, we construct a benchmark for video mirror reflection generation by repurposing four existing video mirror datasets for our task.
The benchmark includes diverse scenes, object configurations, camera motions, and mirror appearances, supporting both model training and systematic evaluation. Experiments demonstrate improved reflection reconstruction quality over representative image-based reflection generation methods and strong video inpainting baselines. Ablation studies show complementary roles of semantic relation learning and transformation alignment, supporting the proposed \emph{what-to-reflect} and \emph{how-to-reflect} decomposition. Our contributions are:
\begin{itemize}
    \item We introduce the novel task of video mirror reflection generation and identify two key challenges: \emph{what} content to reflect and \emph{how} to arrange it within mirrors.
    \item We propose \textbf{MirrorWorld}, a reflection-aware video inpainting framework integrating Semantic Relation Distillation (SRD) and Geometric Transformation Alignment (GTA) to align visible scene content with mirror regions.
    \item We construct the first benchmark for video mirror reflection generation.
    \item Experiments demonstrate that MirrorWorld outperforms representative image reflection generation methods and strong video generation baselines in reflection quality.
\end{itemize}

\section{Related Work}
\label{sec:relate}

\subsection{Video Diffusion Models}
Diffusion models have become a dominant paradigm for video generation. Video Diffusion Models~\cite{vdm} extend image diffusion models to jointly model spatial appearance and temporal dynamics, while VideoLDM~\cite{videoldm} enables high-resolution video generation in the latent space. Recent methods further improve visual quality and scalability through motion modules and diffusion transformers~\cite{animatediff,VideoCrafter2,cogvideox}. Recent work has also investigated physical plausibility in generated videos. VideoPhy~\cite{VideoPhy} and PhyGenBench~\cite{worldsimulator} evaluate physical commonsense, while VLIPP~\cite{VLIPP} and PhysHPC~\cite{physhpc} improve physical fidelity using physical priors and preference alignment. Inspired by REPA~\cite{repa}, VideoREPA~\cite{videorepa} transfers relational representations from video understanding to generation models. However, these methods focus on motion quality, physical plausibility, or representation alignment for video generation. In contrast, mirror reflection generation requires reasoning about relationships between visible scene content and mirror regions. Unlike VideoREPA, which aligns all spatio-temporal token relations, SRD distills only cross-region relations between mirror and visible regions. MirrorWorld addresses this challenge through semantic relation learning and transformation alignment for reflection generation.

\begin{figure*}[!ht] \centering
    \includegraphics[width=1\textwidth]{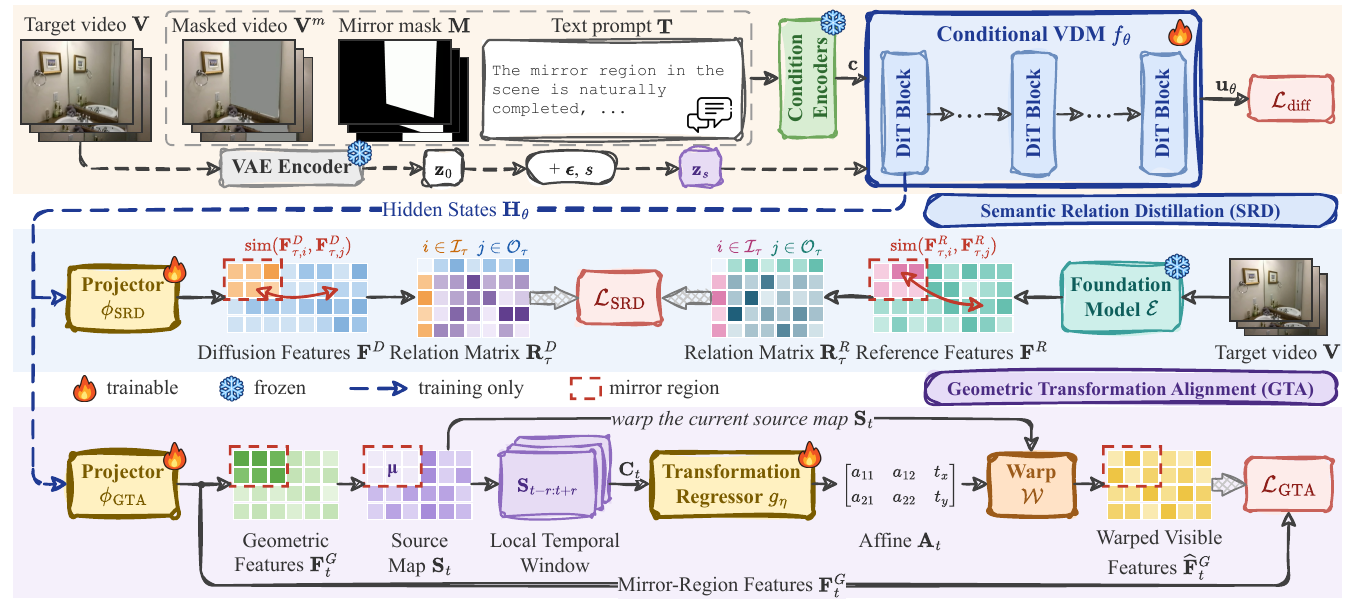}
    \vspace{-6mm}
    \caption{
    Overview of \textbf{MirrorWorld}. Given a masked video, a mirror mask, and
    a text prompt, the conditional VDM reconstructs the missing mirror regions.
    During training, Semantic Relation Distillation (SRD) transfers cross-region
    relations from a frozen visual foundation model to constrain \emph{what}
    visible content should be reflected, while Geometric Transformation Alignment
    (GTA) learns temporally conditioned feature-space transformations to constrain
    \emph{how} this content should be arranged.
    }
    \label{framework}
    \vspace{-2mm}
\end{figure*}

\subsection{Video Inpainting Methods}
Video inpainting reconstructs missing regions from visible frames and text prompts. Early data-driven methods exploit information across frames through optical flow, feature propagation, and transformers~\cite{E2FGVI,FGT,propainter}. Recent generative methods leverage video diffusion priors to improve visual quality and diversity~\cite{AVID,cococo,videopainter,vace,EditCtrl}. However, these methods are not designed to model relationships between mirror regions and the surrounding scene and, when applied to mirrors, may rely on local context and generative priors. In mirror region inpainting, generated content should remain consistent with visible scene content. MirrorWorld instead explicitly models the dependency between visible context and mirror regions.

\subsection{Mirror Understanding and Generation}
Mirror-related vision research has primarily focused on mirror detection and segmentation. Image-based methods identify mirrors using contextual similarity~\cite{pmd}, depth discontinuities~\cite{depthmirror}, and semantic associations~\cite{semanticmirror}. Video-based methods further leverage spatio-temporal relationships~\cite{vmd}, weak supervision~\cite{zoom}, and motion cues~\cite{mmd} to improve mirror localization across frames. While these methods exploit mirror-scene relationships for localization and segmentation, they do not address reflection generation. More recently, MirrorFusion~\cite{mirrorfusion} introduces a diffusion-based mirror reflection generation method using image inpainting and depth conditioning for single-object reflection synthesis. MirrorVerse~\cite{mirrorverse} extends this setting to multi-object configurations. However, existing reflection generation methods remain restricted to image-based and object-centric settings.
Beyond static images, video editing methods such as ROSE~\cite{rose} and Generative Omnimatte~\cite{generativeomnimatte} handle reflections as side effects of object removal or layer decomposition, which erase reflections rather than synthesize mirror content from the visible scene.
Meanwhile, 3D reconstruction methods explicitly model mirrors and reflections using neural radiance fields~\cite{nerfren,msnerf,mirrornerf} or 3D Gaussian splatting~\cite{mirrorgaussian}. Other works exploit observed reflections as constraints to recover scene geometry~\cite{reflect3r}, while Ref-GeNVS~\cite{Ref-GeNVS} adapts multi-view diffusion models to synthesize novel views of mirror scenes. However, all these methods rely on reflections that are already present in the input views, whereas our task synthesizes missing reflections within masked mirror regions.
Furthermore, existing video mirror datasets~\cite{vmd,zoom,mmd,DVMD-D} are designed exclusively for detection and segmentation rather than reflection reconstruction. In contrast, MirrorWorld addresses reflection generation in videos by leveraging relationships between visible content and mirror regions, handling scene-level reflections in real-world environments, and establishing the first benchmark for systematic evaluation.

\section{Method}
\label{sec:method}

\subsection{Overview}

Our goal is to develop a video mirror reflection generation framework,
denoted as \textbf{MirrorWorld}, which reconstructs reflection-consistent
mirror content while preserving the visible non-mirror regions. Formally,
given an input video $\mathbf{V}^{m}$ with its mirror regions masked, the
corresponding binary mirror mask $\mathbf{M}$, and a text prompt $\mathbf{T}$,
MirrorWorld generates video
$
    \widehat{\mathbf{V}}
    =
    \mathcal{G}(\mathbf{V}^{m},\mathbf{M},\mathbf{T}),
$
where $\mathcal{G}$ denotes the conditional video generation model. The
generated video $\widehat{\mathbf{V}}$ is expected to preserve the observed
scene content while reconstructing mirror regions that remain semantically
compatible and geometrically consistent with the visible scene.

To achieve this, MirrorWorld integrates two complementary components: (1) \textit{Semantic Relation Distillation (SRD)}, which transfers relational knowledge from a frozen visual foundation model to establish semantic correspondences between mirror regions and reflection-relevant content in the visible scene, and (2) \textit{Geometric Transformation Alignment (GTA)},
which aligns visible-scene and mirror-region features through a learned feature-space transformation, providing a coarse geometric prior on how the associated content is arranged within the mirror.
Both components are used only during training. Figure~\ref{framework} illustrates the overall architecture.

\subsection{Semantic Relation Distillation (SRD)}

Semantic Relation Distillation (SRD) aims to establish semantic correspondences between mirror regions and the visible scene. A plausible reflection should contain elements that are semantically related to the surrounding visible content. However, such scene-to-mirror semantic relations are not explicitly encoded in the pretrained VDMs. Hence, we distill semantic relational knowledge from a frozen visual foundation model~\cite{videomae} into the diffusion representation.

We employ a frozen visual foundation model~\cite{videomae} $\mathcal{E}$ to extract reference semantic features $\mathbf{F}^{R}=\mathcal{E}(\mathbf{V})$ from the target video $\mathbf{V}$.
Given the noisy video latent $\mathbf{z_s}$ at diffusion timestep $s$, we extract a hidden representation
$
\mathbf{H_\theta}=f_\theta(\mathbf{z_s,s,\mathbf{V}^{m},\mathbf{M},\mathbf{T}})
$
from the VDM $f_\theta$, which is then passed through a lightweight MLP projection head $\phi_{\mathrm{SRD}}$ to obtain the diffusion features
$
\mathbf{F}^{D} =
\phi_{\mathrm{SRD}}(\mathbf{H}_{\theta}).
$
The diffusion features and mirror mask are resampled onto the reference token grid.

For each valid temporal slice $\tau$, let $\mathcal{I}_{\tau}$ and
$\mathcal{O}_{\tau}$ denote the sampled token indices inside and outside the
mirror, respectively. We measure the relation between two feature vectors using cosine similarity:
$\operatorname{sim}(\mathbf{a},\mathbf{b})
=\mathbf{a}^{\top}\mathbf{b}/
(\|\mathbf{a}\|_{2}\|\mathbf{b}\|_{2})$.
The scene-to-mirror semantic relations obtained from the reference features $\mathbf{F}^{R}$ and diffusion features are defined as:
\begin{equation}
\begin{aligned}
\mathbf{R}^{R}_{\tau}(i,j)
&=
\operatorname{sim}
\left(
    \mathbf{F}^{R}_{\tau,i},
    \mathbf{F}^{R}_{\tau,j}
\right),\\
\mathbf{R}^{D}_{\tau}(i,j)
&=
\operatorname{sim}
\left(
    \mathbf{F}^{D}_{\tau,i},
        \mathbf{F}^{D}_{\tau,j}
\right).
\end{aligned}
\label{eq:relation}
\end{equation}
The SRD objective aligns the scene-to-mirror semantic relation as
\begin{equation}
    \mathcal{L}_{\mathrm{SRD}}=\frac{1}{|\mathcal{V}_{S}|}
    \sum_{\tau\in\mathcal{V}_{S}}
    \frac{1}{
        |\mathcal{I}_{\tau}||\mathcal{O}_{\tau}|
    }
    \sum_{\substack{i\in\mathcal{I}_{\tau}\\
                    j\in\mathcal{O}_{\tau}}}
    \left|
        \mathbf{R}^{R}_{\tau}(i,j)
        -
        \mathbf{R}^{D}_{\tau}(i,j)
    \right|,
\label{eq:srd}
\end{equation}
where $\mathcal{V}_{S}$ contains the temporal slices with sufficient tokens in both regions.
With $\mathcal{E}$ frozen and gradients stopped on visible features $\mathbf{F}_{\tau,j}^{D}$ ($j \in \mathcal{O}_\tau$), the visible scene serves as a fixed anchor and $\mathcal{L}_{\text{SRD}}$ shapes only the mirror-region features. Computing $\mathbf{R}^{R}$ on the ground-truth video further enables teacher relations to identify which visible objects are reflected.
In contrast to VideoREPA~\cite{videorepa}, which enforces global spatio-temporal relation matching across all video tokens for general synthesis, SRD formulates reflection generation as asymmetric cross-region reasoning within each frame. Rather than co-adapting all token pairs, SRD couples masked mirror tokens exclusively with visible-scene tokens and anchors the latter via gradient detachment, using the observed scene as fixed contextual cues to infer mirror semantics.

\subsection{Geometric Transformation Alignment (GTA)}

SRD identifies semantic relationships between mirror regions and visible
scene content, but it does not explicitly determine how the associated
content should be spatially arranged within the reflection. We therefore
introduce Geometric Transformation Alignment (GTA) to learn a
feature-space alignment between the visible content and the mirror region.

We obtain the geometric features
$\mathbf{F}^{G}=\phi_{\mathrm{GTA}}(\mathbf{H}_{\theta})$
using a separate trainable MLP projection head
$\phi_{\mathrm{GTA}}$.
For each temporal index $t$, we construct a source map containing only visible
scene information:
\begin{equation}
    \mathbf{S}_{t}(\mathbf{q})
    =
    \begin{cases}

            \mathbf{F}^{G}_{t}(\mathbf{q})
        ,
        & \mathbf{q}\in\mathcal{O}_{t},\\

            \tfrac{1}{|\mathcal{O}_{t}|}
            \sum_{\mathbf{p}\in\mathcal{O}_{t}}
            \mathbf{F}^{G}_{t}(\mathbf{p})
        ,
        & \mathbf{q}\in\mathcal{I}_{t}.
    \end{cases}
\label{eq:gta_source}
\end{equation}
where $\mathcal{I}_{t}$ and $\mathcal{O}_{t}$ denote the sets of locations
inside and outside the mirror, respectively.
Replacing mirror-region features with the mean visible feature avoids a trivial identity solution, while stopping gradients through $\mathbf{S}_t$ ensures $\mathcal{L}_{\text{GTA}}$ updates only the mirror features and regressor.

Estimating the transformation independently at each temporal step may produce
unstable geometric mappings. GTA therefore aggregates a local temporal window
of $K=2r+1$ feature steps, where $r$ is the temporal radius. Specifically, we
compute the local context as
\begin{equation}
    \mathbf{C}_{t}=\frac{1}{K}\sum_{k=-r}^{r}
    \mathbf{S}_{\operatorname{clip}(t+k)},
\end{equation}
where $\operatorname{clip}(\cdot)$ replicates boundary indices. The window is
defined over the temporal grid of the video features.

A shared transformation regressor $g_{\eta}$ predicts an affine matrix from
the temporal context
$
    \mathbf{A}_{t}
    =
    g_{\eta}(\mathbf{C}_{t}).
$
The regressor predicts an affine transformation initialized from the identity mapping.
The predicted transformation is then applied to the source feature at the
current temporal index to obtain the warped visible features
\begin{equation}
    \widehat{\mathbf{F}}^{G}_{t}
    =
    \mathcal{W}
    \left(
        \mathbf{S}_{t},
        \mathbf{A}_{t}
    \right),
\end{equation}
where $\mathcal{W}$ denotes differentiable bilinear warping. Thus, the
$K$-step context serves only to estimate $\mathbf{A}_{t}$, while this transformation is applied solely to the current source feature map $\mathbf{S}_{t}$.
GTA measures the cosine distance between the warped visible features and the corresponding mirror-region features:
\begin{equation}
\mathcal{L}_{\mathrm{GTA}}
    =
    \frac{1}{|\mathcal{V}_{G}|}
    \sum_{t\in\mathcal{V}_{G}}
    \frac{1}{|\mathcal{I}_{t}|}
    \sum_{\mathbf{q}\in\mathcal{I}_{t}}
    \left[
        1-
        \operatorname{sim}
        \left(
            \widehat{\mathbf{F}}^{G}_{t}(\mathbf{q}),
            \mathbf{F}^{G}_{t}(\mathbf{q})
        \right)
    \right].
\label{eq:gta_loss}
\end{equation}
Here, $\mathcal{V}_{G}$ contains the temporal indices with sufficient
mirror-region and outside-region support. This objective encourages the
transformed visible features to match the mirror-region representation.

\subsection{Training Objective}

Given a target video $\mathbf{V}$, the video VAE encodes it into a clean
latent $\mathbf{z}_{0}$. We sample a diffusion timestep $s$ and Gaussian noise
$\boldsymbol{\epsilon}\sim\mathcal{N}(\mathbf{0},\mathbf{I})$ to construct
the noisy latent
$\mathbf{z}_{s}=q_{s}(\mathbf{z}_{0},\boldsymbol{\epsilon})$.
Let $\mathbf{c}=(\mathbf{V}^{m},\mathbf{M},\mathbf{T})$ denote the masked
video, mirror mask, and text condition. The base diffusion objective is
\begin{equation}
    \mathcal{L}_{\mathrm{diff}}
    =
    \mathbb{E}_{\mathbf{z}_{0},s,\boldsymbol{\epsilon}}
    \left[
        w(s)\,
        \operatorname{MSE}
        \left(
            \mathbf{u}_{\theta}(\mathbf{z}_{s},s,\mathbf{c}),
            \mathbf{u}_{s}^{*}
        \right)
    \right],
\label{eq:diffusion_loss}
\end{equation}
where $\mathbf{u}_{\theta}$ is the model prediction,
$\mathbf{u}_{s}^{*}$ is the scheduler-defined training target, and $w(s)$
is the timestep-dependent weighting function.
The final objective combines the diffusion generation loss with SRD and GTA:
\begin{equation}
    \mathcal{L}
    =
    \mathcal{L}_{\mathrm{diff}}
    +
    \lambda_{\mathrm{SRD}}\mathcal{L}_{\mathrm{SRD}}
    +
    \lambda_{\mathrm{GTA}}\mathcal{L}_{\mathrm{GTA}},
\label{eq:overall_objective}
\end{equation}
where $\lambda_\mathrm{SRD}$ and $\lambda_\mathrm{GTA}$ control the strength of the two objectives. This strategy enables the VDM to learn \emph{what} visible scene content is associated with the mirror region from SRD and \emph{how} this content is spatially transformed within the reflection in GTA. During inference, we discard the SRD and GTA heads and sample from Gaussian noise conditioned solely on $(\mathbf{V}^{m},\mathbf{M},\mathbf{T})$, introducing no additional inference overhead.

\subsection{Benchmark}

We construct a video mirror reflection generation benchmark from four existing video mirror segmentation and detection datasets: VMD-D~\cite{vmd}, ZOOM~\cite{zoom}, MMD~\cite{mmd}, and DVMD-D~\cite{DVMD-D}. These datasets provide real-world videos together with annotated mirror masks, covering diverse scenes, camera motions, and mirror appearances. By integrating the four datasets into a shared evaluation environment, our benchmark defines a unified reflection-aware video reconstruction task, in which all methods receive the same masked videos, mirror masks, and text prompts.

Since the original videos vary substantially in length, we divide them into clips of at most 49 frames and discard clips shorter than 10 frames to satisfy the input-length constraints of VDMs. This process produces 1,242 video clips. We perform the train-test split at the source-video level, ensuring that the clips from the same source video never appear in both the training and testing sets. This results in 1,142 training clips and 100 testing clips. For each clip, the original video serves as the target, while the conditional input is constructed by masking the annotated mirror regions.

\section{Experiments}
\label{sec:experiments}

\subsection{Evaluation Protocol}

\noindent\textbf{Implementation details.}
Our model is built upon Wan2.1-VACE-14B~\cite{wan, vace}, which is fine-tuned using LoRA~\cite{lora} with rank 32.
For SRD, we employ the frozen VideoMAEv2-Base~\cite{videomae} as the visual foundation model.
The GTA transformation estimator uses a local window of $K=5$ to predict the affine transformation.
$\mathbf{H}_\theta$ represents the input hidden states to the 20th block of the 40-block DiT backbone.
The weights of the SRD and GTA objectives are set to $\lambda_{\mathrm{SRD}}=0.05$ and $\lambda_{\mathrm{GTA}}=0.01$, respectively.
We train MirrorWorld on 4 NVIDIA A100 80GB GPUs with a global batch size of 4, using the AdamW~\cite{adamw} optimizer with a learning rate of \(10^{-4}\).

\noindent\textbf{Baselines.}
We compare MirrorWorld with two categories of baseline methods. Image mirror reflection generation methods include MirrorFusion~\cite{mirrorfusion} and MirrorVerse~\cite{mirrorverse}, which are applied independently to each video frame.
Video inpainting models include VideoPainter~\cite{videopainter} and VACE~\cite{vace}. VACE adopts Wan2.1-14B~\cite{wan} as the base model.
To ensure a fair comparison, all baseline methods are fine-tuned on our training set using their official implementations. During evaluation, all methods use the same masked videos, mirror masks, and text prompts.

\noindent\textbf{Metrics.}
Within mirror regions, we measure pixel and perceptual fidelity using PSNR, SSIM, and LPIPS, temporal stability using flow warping error $E_{\text{warp}}$~\cite{ewarp}, semantic consistency using DINO based on DINOv3~\cite{dinov3} features, and shading error $\mathrm{SH}_{\mathrm{err}}$ from MoGe-2~\cite{moge2} normals, alongside full-frame FVD for overall video quality.
We also use GPT-6.1 Sol as a VLM judge to evaluate the masked input and generated videos, independently scoring reflection geometry (Geo.), semantic consistency (Sem.), temporal consistency (Temp.), and visual quality (Vis.) on a 0-100 scale.
Each method is evaluated at its native output resolution by resizing the ground truth and mask via bicubic and nearest-neighbor interpolation, respectively.

\begin{table*}[t]
\centering
\resizebox{\linewidth}{!}{
\begin{tabular}{lccccccccccc}
\toprule
& \multicolumn{3}{c}{\textbf{Reflection Fidelity}}
& \multicolumn{2}{c}{\textbf{Video Quality}}
& \multicolumn{2}{c}{\textbf{Consistency}}
& \multicolumn{4}{c}{\textbf{VLM Judge}} \\
\cmidrule(lr){2-4} \cmidrule(lr){5-6} \cmidrule(lr){7-8} \cmidrule(lr){9-12}
\textbf{Method}
& \textbf{PSNR} $\uparrow$
& \textbf{SSIM} $\uparrow$
& \textbf{LPIPS} $\downarrow$
& \textbf{FVD} $\downarrow$
& \textbf{\boldmath$E_{\mathrm{warp}}$} $\downarrow$
& \textbf{DINO} $\uparrow$
& \textbf{\boldmath$\mathrm{SH}_{\mathrm{err}}$} $\downarrow$
& \textbf{Geo.} $\uparrow$
& \textbf{Sem.} $\uparrow$
& \textbf{Temp.} $\uparrow$
& \textbf{Vis.} $\uparrow$ \\
\midrule
\rowcolor{gray!10}
\multicolumn{12}{l}{\textit{\textbf{Image Reflection Generation Models}}} \\
MirrorFusion
~\cite{mirrorfusion}
& 9.508
& 0.293
& 0.699
& 513.062
& 0.192
& 0.687
& 64.3
& 14.0
& 15.7
& 7.8
& 31.0 \\

MirrorVerse
~\cite{mirrorverse}
& 9.666
& 0.312
& 0.680
& 416.146
& 0.188
& 0.719
& 68.4
& 24.4
& 27.3
& 14.7
& 35.6 \\
\midrule
\rowcolor{gray!10}
\multicolumn{12}{l}{\textit{\textbf{Video Generation Models}}}\\
VideoPainter
~\cite{videopainter}
& 11.282
& 0.399
& 0.606
& 229.558
& 0.076
& 0.765
& 44.7
& 59.6
& 60.4
& 58.5
& 51.2 \\

VACE
~\cite{vace}
& \underline{13.537}
& \underline{0.489}
& \underline{0.493}
& \underline{191.617}
& \underline{0.026}
& \underline{0.819}
& \underline{39.0}
& \underline{78.9}
& \underline{79.9}
& \underline{84.0}
& \underline{72.1} \\

\textbf{Ours}
& \textbf{14.006}
& \textbf{0.504}
& \textbf{0.488}
& \textbf{184.868}
& \textbf{0.025}
& \textbf{0.820}
& \textbf{36.6}
& \textbf{79.6}
& \textbf{82.2}
& \textbf{84.7}
& \textbf{72.9} \\
\bottomrule
\end{tabular}
}
\vspace{-2mm}
\caption{
\textbf{Quantitative comparison with baseline methods.}
VLM Judge reports blind scores on a 0-100 scale for reflection geometry (Geo.), semantic consistency (Sem.), temporal consistency (Temp.), and visual quality (Vis.).
The best and second-best results are highlighted in \textbf{bold} and \underline{underlined}.
}
\label{tab:main_results}
\vspace{-4mm}
\end{table*}

\subsection{Comparison with Baseline Methods}
\label{sec:main_results}

\noindent\textbf{Quantitative evaluation.}
Table~\ref{tab:main_results} compares MirrorWorld with representative image
mirror reflection generation and video inpainting methods. MirrorWorld
achieves the strongest overall performance across both mirror-region
reconstruction and video-level evaluation.

Compared with the VACE backbone, MirrorWorld improves pixel, structural, and perceptual fidelity, while obtaining the highest DINO similarity and lowest shading error, reflecting closer alignment with target content and shading. It also achieves the best FVD and lowest $E_{\text{warp}}$, demonstrating superior video realism and temporal stability. The blind VLM judge rates MirrorWorld highest on all four criteria, most notably on semantic consistency. These gains confirm the benefit of reflection-specific semantic and geometric supervision beyond conventional video inpainting.
Video-based methods generally outperform image-based approaches. Since MirrorFusion and MirrorVerse process frames independently without modeling cross-frame reflection consistency, they show weaker video-level performance and achieve the lowest VLM temporal consistency.

\begin{table}[t]
    \centering
    \resizebox{\linewidth}{!}{
    \begin{tabular}{lccccc}
        \toprule
        & \multicolumn{3}{c}{\textbf{Reflection Fidelity}}
        & \multicolumn{2}{c}{\textbf{Video Quality}} \\
        \cmidrule(lr){2-4} \cmidrule(lr){5-6}
        \textbf{Method}
        & \textbf{PSNR} $\uparrow$
        & \textbf{SSIM} $\uparrow$
        & \textbf{LPIPS} $\downarrow$
        & \textbf{FVD} $\downarrow$
        & \textbf{\boldmath$E_{\mathrm{warp}}$} $\downarrow$ \\
        \midrule
        \rowcolor{gray!10}
        \multicolumn{6}{l}{\textit{\textbf{Image Reflection Generation Models}}} \\
        MirrorFusion~\cite{mirrorfusion} & 10.811 & 0.288 & 0.665 & 910.785 & 0.184 \\
        MirrorVerse~\cite{mirrorverse}   & 10.512 & 0.304 & 0.659 & 813.111 & 0.174 \\
        \midrule
        \rowcolor{gray!10}
        \multicolumn{6}{l}{\textit{\textbf{Video Generation Models}}} \\
        VideoPainter~\cite{videopainter} & 12.041 & 0.380 & 0.610 & 482.674 & 0.059 \\
        VACE~\cite{vace}                 & \underline{13.170} & \underline{0.448} & \underline{0.502} & \underline{339.561} & \underline{0.020} \\
        \textbf{Ours}                    & \textbf{13.879} & \textbf{0.469} & \textbf{0.498} & \textbf{321.824} & \textbf{0.018} \\
        \bottomrule
    \end{tabular}
    }
    \vspace{-2mm}
    \caption{\textbf{Quantitative comparison with baseline methods on multi-mirror scenes.} The best and second-best results are highlighted in \textbf{bold} and \underline{underlined}.}
    \label{tab:multi_mirror}
    \vspace{-4mm}
\end{table}

\begin{figure*}[!t] \centering
    \includegraphics[width=1\textwidth]{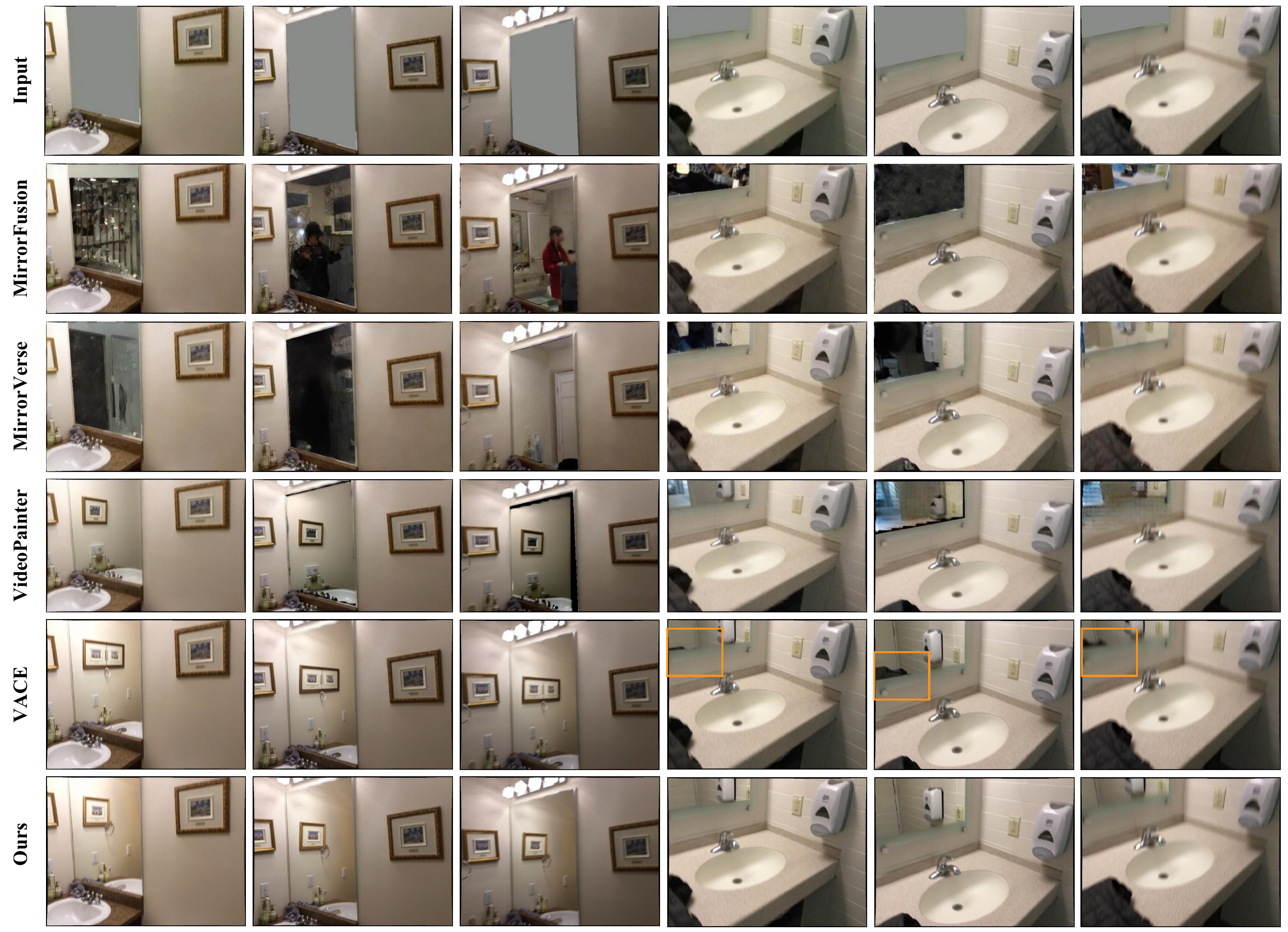}
    \vspace{-6mm}
    \caption{\textbf{Qualitative comparisons on video mirror reflection
    generation.} Existing methods introduce unrelated content or place reflected
    objects in implausible regions, whereas MirrorWorld generates reflections
    consistent with the visible scene and its spatial arrangement. Orange boxes
    highlight incorrectly placed reflections.}
    \label{qualitative}
    \vspace{-6mm}
\end{figure*}

\noindent\textbf{Qualitative evaluation.}
Figure~\ref{qualitative} presents qualitative comparisons on two test videos.
The image-based methods suffer from semantic and temporal inconsistencies.
In the first example, MirrorFusion generates unrelated shelves and people
inside the mirror, while MirrorVerse produces dark or structurally unrelated
content. These reflections also change substantially across frames.
VideoPainter generates more coherent local appearance, but the reflected
content remains weakly aligned with the surrounding scene and contains
inconsistent boundaries and spatial arrangements.
The video-based VACE baseline better preserves the overall scene structure,
yet still fails to maintain the scene-to-mirror relationship. In the first
example, it duplicates the wall painting and places two framed objects inside
the mirror. In the second example, VACE incorrectly generates clothing
reflections in the regions highlighted by the orange boxes, although the
visible clothing positions indicate that no such reflections should appear
there. In contrast, MirrorWorld avoids such duplicated and misplaced reflections, validating our design in which SRD identifies relevant visible semantics and GTA governs their spatial arrangement inside the mirror.

\noindent\textbf{Evaluation on multi-mirror scenes.}
GTA predicts a single frame-level transformation shared across all mirrors within a frame. To examine whether this shared parameterization limits performance in multi-mirror scenes, Table~\ref{tab:multi_mirror} evaluates the 23 test clips containing multiple mirrors. MirrorWorld consistently outperforms VACE, indicating that sharing a single transformation does not hinder multi-mirror synthesis on our benchmark. While metrics are computed over the union of mirror masks within each frame, extending GTA to predict instance-specific transformations for non-coplanar mirrors remains a promising direction for future work.

\begin{figure}[t] \centering
    \includegraphics[width=\columnwidth]{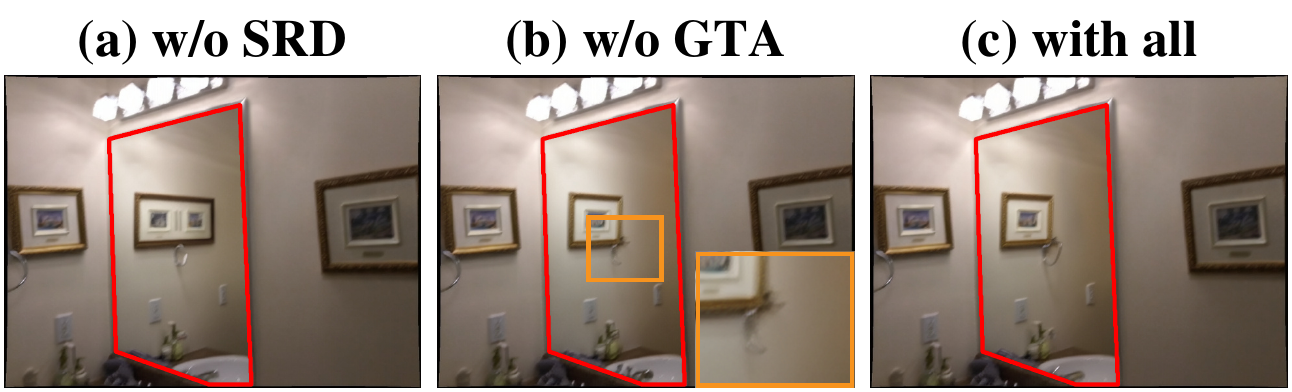}
    \vspace{-6mm}
    \caption{\textbf{Visual ablation of key components.}
    Red lines outline the input mirror mask.}
    \label{fig:component_ablation}
    \vspace{-4mm}
\end{figure}

\begin{table}[t]
\centering
\resizebox{\linewidth}{!}{
\begin{tabular}{cccccc}
\toprule
\textbf{SRD}
& \textbf{GTA}
& \textbf{PSNR} $\uparrow$
& \textbf{SSIM} $\uparrow$
& \textbf{LPIPS} $\downarrow$
& \textbf{FVD} $\downarrow$ \\
\midrule

&
& 13.537
& 0.489
& 0.493
& 191.617 \\

\checkmark
&
& 13.551
& 0.500
& 0.489
& 185.708 \\

& \checkmark
& 13.542
& 0.495
& 0.490
& \textbf{174.444} \\

\checkmark
& \checkmark
& \textbf{14.006}
& \textbf{0.504}
& \textbf{0.488}
& 184.868 \\
\bottomrule
\end{tabular}
}
\vspace{-2mm}
\caption{
Ablation study on key components.
}
\vspace{-4mm}
\label{tab:module_ablation}

\end{table}

\subsection{Ablation Study}
\label{sec:ablation}

\noindent\textbf{Key components.}
We conduct ablation studies with VACE~\cite{vace} as the baseline to examine the contributions of SRD and GTA.
As shown in Table~\ref{tab:module_ablation}, using either SRD or GTA alone
provides limited improvements in mirror-region reconstruction.
SRD models \emph{what} visible content to reflect through cross-region semantic association without constraining spatial layout, whereas GTA models \emph{how} to geometrically arrange that content inside the mirror without explicit semantic supervision.
Each component alone therefore addresses only one aspect of reflection generation.
Notably, GTA alone achieves the lowest FVD among all variants by leveraging local temporal context to provide consistent spatial guidance across neighboring frames, thereby improving video-level distribution quality.
Combining SRD and GTA produces the strongest mirror-region reconstruction, supporting their complementary roles in jointly constraining the semantic content and spatial organization of the generated reflection.

Figure~\ref{fig:component_ablation} further illustrates the complementary
roles of SRD and GTA. Without SRD, the model generates two adjacent framed
artworks inside the mirror, producing semantically incorrect reflections. Without
GTA, it reconstructs only part of the metal ring, as highlighted by the
orange box. Since only half of the ring is visible outside the mirror, this
incomplete reflection indicates that the model merely copies the visible
fragment rather than learning its geometric transformation. In contrast, the full
model reconstructs the complete ring with coherent spatial arrangement, confirming that SRD captures \emph{what} visible content should be reflected while GTA governs \emph{how} it is geometrically arranged in the mirror.

\begin{table}[t]
    \centering
\resizebox{\linewidth}{!}
{
    \begin{tabular}{lcccc}
        \toprule
        \textbf{Model}
        & \textbf{PSNR} $\uparrow$
        & \textbf{SSIM} $\uparrow$
        & \textbf{LPIPS} $\downarrow$
        & \textbf{FVD} $\downarrow$ \\
        \midrule
        DINOv3~\cite{dinov3}
        & 13.503
        & 0.494
        & 0.494
        & \textbf{183.304} \\
        \textbf{VideoMAEv2~\cite{videomae} (Ours)}
        & \textbf{14.006}
        & \textbf{0.504}
        & \textbf{0.488}
        & 184.868 \\
        \bottomrule
    \end{tabular}
    }
    \vspace{-2mm}
        \caption{Ablation on the visual foundation model in SRD.}
    \label{tab:vfm_ablation}
    \vspace{-4mm}
\end{table}

\begin{table}[t]
\centering
\resizebox{\linewidth}{!}{
\begin{tabular}{lcccc}
\toprule
& \textbf{Prec.} $\uparrow$ & \textbf{Rec.} $\uparrow$ & \textbf{F1} $\uparrow$ & \textbf{Halluc.} $\downarrow$ \\
\midrule
VACE~\cite{vace}               & 0.594 & 0.629 & 0.583 & 0.266 \\
+ VideoREPA~\cite{videorepa}   & 0.579 & 0.625 & 0.572 & 0.291 \\
\textbf{+ SRD (Ours)}                          & \textbf{0.610} & \textbf{0.635} & \textbf{0.597} & \textbf{0.260} \\
\bottomrule
\end{tabular}}
\vspace{-2mm}
\caption{\textbf{Ablation on semantic relational objectives in SRD.} Prec., Rec., and F1 denote semantic precision, recall, and F1 score over mirror-region categories, while Halluc.\ denotes the fraction of generated categories absent from both the ground-truth mirror and visible scene.}
\label{tab:relation_placement}
\vspace{-4mm}
\end{table}

\begin{table}[t]
    \centering
    \resizebox{\linewidth}{!}
    {
    \begin{tabular}{lcccc}
        \toprule
        \textbf{Transformation}
        & \textbf{PSNR} $\uparrow$
        & \textbf{SSIM} $\uparrow$
        & \textbf{LPIPS} $\downarrow$
        & \textbf{FVD} $\downarrow$ \\
        \midrule
        Flip
        & 13.802
        & 0.501
        & 0.490
        & 189.707 \\
        \textbf{Geometric (Ours)}
        & \textbf{14.006}
        & \textbf{0.504}
        & \textbf{0.488}
        & \textbf{184.868} \\
        \bottomrule
    \end{tabular}
    }
    \vspace{-2mm}
    \caption{Ablation on the transformation strategy in GTA.}
    \label{tab:transform_ablation}
    \vspace{-4mm}
\end{table}

\begin{table}[t]
    \centering
    \resizebox{\linewidth}{!}
    {
    \begin{tabular}{lcccc}
        \toprule
        \textbf{Temporal Context}
        & \textbf{PSNR} $\uparrow$
        & \textbf{SSIM} $\uparrow$
        & \textbf{LPIPS} $\downarrow$
        & \textbf{FVD} $\downarrow$ \\
        \midrule
        $K=1$ (Per-frame)
        & 13.543
        & 0.494
        & 0.499
        & \textbf{175.789} \\
        \textbf{\boldmath$K=5$ (Ours)}
        & \textbf{14.006}
        & \textbf{0.504}
        & \textbf{0.488}
        & 184.868 \\
        $K=7$
        & 13.541
        & 0.495
        & 0.494
        & 181.303 \\
        All frames
        & 13.528
        & 0.497
        & 0.490
        & 189.899 \\
        \bottomrule
    \end{tabular}
    }
    \vspace{-2mm}
    \caption{Ablation on the temporal window size in GTA.}
    \label{tab:temporal_window_ablation}
    \vspace{-4mm}
\end{table}

\noindent\textbf{Visual foundation model in SRD.}
We investigate the choice of visual foundation model in SRD by replacing
VideoMAEv2~\cite{videomae} with DINOv3~\cite{dinov3}. As shown in Table~\ref{tab:vfm_ablation}, VideoMAEv2
outperforms DINOv3 across all three mirror-region reconstruction metrics,
indicating more accurate reflected content. Although DINOv3 obtains a slightly lower
FVD, this minor gain in video-level distribution quality
does not translate into improved mirror-region reconstruction. These results
confirm that VideoMAEv2 provides more suitable semantic relational supervision for SRD.

\noindent\textbf{Semantic relational objective in SRD.}
To isolate our semantic relational objective, we replace SRD with the VideoREPA~\cite{videorepa} relation loss under identical settings and evaluate mirror-region semantic categories segmented by OneFormer~\cite{oneformer}. As shown in Table~\ref{tab:relation_placement}, applying global relation alignment via VideoREPA degrades semantic accuracy and increases hallucinations relative to VACE~\cite{vace}, as aligning all token pairs dilutes supervision across the visible background and updates visible-scene features symmetrically rather than keeping them fixed. In contrast, by restricting supervision to mirror-visible pairs with gradient-detached scene anchors, SRD consistently outperforms both baselines across precision, recall, F1, and hallucination rate, confirming that asymmetric cross-region supervision yields more faithful reflection semantics.

\noindent\textbf{Transformation strategy in GTA.}
We investigate the transformation strategy in GTA by comparing our geometric
transformation with a learnable flip-based variant that horizontally flips the visible-scene features before applying a
trainable residual MLP for pointwise feature refinement.
Despite these feature corrections, its spatial correspondence remains constrained by horizontal flipping.
As shown in Table~\ref{tab:transform_ablation}, our geometric transformation
outperforms the Flip variant across all three mirror-region reconstruction metrics and
obtains a lower FVD, confirming that refining flipped features alone cannot capture scene-dependent reflection geometry. By estimating spatial transformations from local temporal context, GTA more effectively learns how to arrange visible content within the reflection.

\noindent\textbf{Temporal window size in GTA.}
We investigate the temporal window size in GTA by comparing per-frame
transformation estimation, local windows of varying sizes, and aggregation
over all frames. In Table~\ref{tab:temporal_window_ablation}, $K=5$ achieves the best performance across all three mirror-region reconstruction metrics, indicating more accurate reflected content. While the per-frame setting
obtains the lowest FVD, its weaker reconstruction metrics indicate no improvement in reflection quality.
Moreover, expanding the window to $K=7$ or aggregating all frames brings no further gains, confirming that GTA benefits most from a moderate local context. Single-frame estimation lacks temporal evidence, whereas an overly broad window dilutes the frame-specific geometric cues required for reflection alignment.

\begin{figure}[t] \centering
    \center
    \includegraphics[width=\columnwidth]{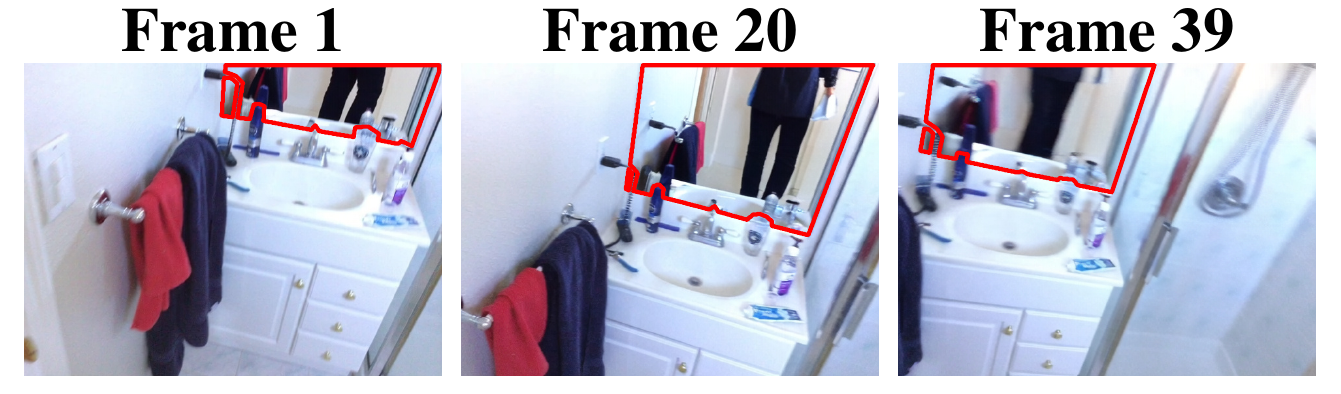}
    \vspace{-6mm}
    \caption{\textbf{Limitation analysis.}
    Red lines outline the mirror mask.}
    \label{fig:failure_case}
    \vspace{-4mm}
\end{figure}

\vspace{-1mm}
\subsection{Limitation}

Since our formulation reconstructs reflections via visible-scene correspondences, SRD and GTA cannot constrain the identity or spatial layout of subjects outside the camera view. When the reflected person stands behind the camera in Figure~\ref{fig:failure_case}, MirrorWorld generates a temporally consistent reflection using learned priors rather than recovering their actual unseen appearance.

\vspace{-1mm}
\section{Conclusion}
\label{sec:conclusion}
\vspace{-1mm}

We presented \textbf{MirrorWorld}, to our knowledge the first framework for mirror reflection generation in videos. MirrorWorld decomposes reflection generation into two
complementary questions: \emph{what} content should be reflected and
\emph{how} it should be spatially arranged. Specifically, SRD transfers semantic relations
from a frozen visual foundation model, while GTA aligns visible scene features
with mirror-region representations through learned geometric transformations.
We also constructed a unified video mirror reflection generation benchmark
from four real-world mirror datasets. Experiments show that MirrorWorld significantly improves
reflection reconstruction and video-level quality over representative
image-based reflection generation and video inpainting methods, while ablation
studies validate the complementary roles of SRD and GTA. While the current framework focuses on feature-space transformations, modeling explicit reflection geometry and content outside the field of view are important directions for future work.

{
    \small
    \bibliographystyle{ieeenat_fullname}
    \bibliography{reference}
}

\clearpage
\maketitlesupplementary

\setcounter{figure}{0}
\renewcommand{\thefigure}{\Alph{figure}}

\setcounter{table}{0}
\renewcommand{\thetable}{\Alph{table}}

\setcounter{section}{0}
\renewcommand{\thesection}{\Alph{section}}

\newcommand{\mainsec}[1]{#1 of the main paper}

\renewcommand{\theHsection}{supp.\Alph{section}}
\renewcommand{\theHfigure}{supp.\Alph{figure}}
\renewcommand{\theHtable}{supp.\Alph{table}}

\begin{center}
    {\Large\textbf{Overview}}
\end{center}

\noindent In this supplementary material, we provide additional analyses, technical details, and qualitative results:
\begin{itemize}
    \setlength{\itemsep}{0pt}
    \item \textbf{Sec.~\ref{sec:supp_asym}: Observed-missing asymmetry in reflection reconstruction.} Visible features as fixed context for the missing mirror content, examined through the visible-side stop-gradient ablation, selective recovery analysis, and GTA evaluation across relation objectives.
    \item \textbf{Sec.~\ref{sec:supp_ablation}: Additional ablation study.} Backbone generalization, loss weights, token sharing, and token sampling.
    \item \textbf{Sec.~\ref{sec:supp_experiments}: Additional experiments and analyses.} Source-dataset results and a training-free geometric baseline.
    \item \textbf{Sec.~\ref{sec:benchmark_statistics}: Benchmark statistics.} Source datasets, resolutions, and the source-level split.
    \item \textbf{Sec.~\ref{sec:supp_formulation}: Formulation of MirrorWorld.} Backbone preliminaries and the full formulations of SRD and GTA.
    \item \textbf{Sec.~\ref{sec:supp_impl}: Implementation details.} Text prompt, algorithms, training settings, and baseline details.
    \item \textbf{Sec.~\ref{sec:metric_definitions}: Metric definitions.} All evaluation metrics, including the VLM-as-a-Judge protocol.
    \item \textbf{Sec.~\ref{sec:supp_qual}: Additional qualitative results.} Comparisons with baselines, single-mirror and multi-mirror results.
\end{itemize}

\section{Observed-Missing Asymmetry in Reflection Reconstruction}
\label{sec:supp_asym}

In reflection reconstruction, the visible scene is observed as input, whereas the content inside the mirror is missing and must be inferred from it. Both objectives of MirrorWorld exploit this asymmetry: SRD supervises only cross-region relations between mirror and visible tokens, while GTA aligns mirror features with spatially transformed visible features. In both objectives, visible features serve as fixed contextual evidence via a stop-gradient. We validate this design through three analyses. First, Sec.~\ref{sec:supp_detach} ablates the visible-side stop-gradient in SRD. Second, Sec.~\ref{sec:supp_selective} examines whether SRD selectively recovers valid reflected content rather than indiscriminately copying visible objects into the mirror. Finally, Sec.~\ref{sec:supp_videorepa} pairs GTA with two relation objectives to evaluate its effect on mirror-region reconstruction.

\subsection{Gradient Asymmetry Detachment}
\label{sec:supp_detach}

SRD applies a stop-gradient to the visible endpoint of each mirror-visible pair. To isolate its effect, we train a variant without GTA that allows direct gradients through both endpoints while keeping all other settings fixed. As shown in Table~\ref{tab:supp_detach}, detaching the visible endpoint improves both mirror reconstruction and background preservation. Without the stop-gradient, mirror PSNR, mirror SSIM, and background PSNR all drop, with both PSNR scores falling below the fine-tuned VACE~\cite{vace} baseline. With the stop-gradient, visible features act as fixed anchors, forcing the relation loss to optimize solely by updating mirror features while the background is supervised exclusively by the generation loss. Conversely, allowing gradients through the visible side lets the relation loss pull visible features toward noisy mirror predictions, which interferes with background reconstruction and causes the relation targets for mirror tokens to drift during training. While shared network parameters can still update visible features, the stop-gradient blocks direct SRD gradients from propagating through the visible endpoint.

\begin{table*}[h]
    \centering
    \resizebox{0.8\linewidth}{!}
    {
    \begin{tabular}{lcccccc}
        \toprule
        & \multicolumn{3}{c}{\textbf{Reflection Fidelity}}
        & \multicolumn{3}{c}{\textbf{Background Preservation}} \\
        \cmidrule(lr){2-4} \cmidrule(lr){5-7}
        \textbf{Method}
        & \textbf{PSNR} $\uparrow$
        & \textbf{SSIM} $\uparrow$
        & \textbf{LPIPS} $\downarrow$
        & \textbf{PSNR} $\uparrow$
        & \textbf{SSIM} $\uparrow$
        & \textbf{LPIPS} $\downarrow$ \\
        \midrule
        VACE~\cite{vace}
        & 13.537 & 0.489 & 0.493 & 32.921 & \textbf{0.929} & 0.111 \\
        + SRD w/o detachment
        & 13.366 & 0.490 & \textbf{0.489} & 32.865 & \textbf{0.929} & \textbf{0.110} \\
        \textbf{+ SRD (Ours)}
        & \textbf{13.551} & \textbf{0.500} & \textbf{0.489} & \textbf{33.059} & \textbf{0.929} & 0.111 \\
        \bottomrule
    \end{tabular}
    }
    \caption{Ablation on visible-side detachment in SRD without GTA. Reflection fidelity and background preservation are evaluated inside and outside the mirror, respectively. Best results are in \textbf{bold}.}
    \label{tab:supp_detach}
\end{table*}

\subsection{Selective Recovery of Reflection-Relevant Content}
\label{sec:supp_selective}

Since a mirror reflects only part of the visible scene, a model can inflate semantic overlap by copying visible objects absent from the target reflection. To separate valid reflection recovery from spurious copying, we decompose the generated mirror categories $Q$ that also appear in the visible scene $O$ into two components:
\begin{equation}
\frac{|Q\cap O|}{|Q|}
=\underbrace{\frac{|Q\cap O\cap G|}{|Q|}}_{\mathrm{Vis.\,Corr.}}
+\underbrace{\frac{|Q\cap(O\setminus G)|}{|Q|}}_{\mathrm{Copy}},
\end{equation}
where $G$ is the ground-truth mirror category set, $\mathrm{Vis.\,Corr.}$ measures the fraction of generated categories shared by the visible scene and ground-truth reflection, and $\mathrm{Copy}$ measures those present in the visible scene but absent from the ground truth. Since OneFormer~\cite{oneformer} may miss small reflected objects, we also report a strict rate $\mathrm{Copy}^{\ast}$ counting only categories with zero detected pixels in the ground-truth mirror region. Following Table~\ref{tab:relation_placement} of the main paper, we evaluate all relational objectives without GTA.

In Table~\ref{tab:supp_selective}, SRD recovers a higher fraction of visible-scene categories in the mirror than VideoREPA~\cite{videorepa}, increasing from 0.582 to 0.618. This gain comes almost entirely from $\mathrm{Vis.\,Corr.}$, which increases while $\mathrm{Copy}$ remains unchanged. Both $\mathrm{Copy}$ and $\mathrm{Copy}^{\ast}$ also remain lower than those of VACE~\cite{vace}, and the $\mathrm{Vis.\,Corr.}$ advantage persists across presence thresholds of 0.25\%, 1\%, and 3\%.

\begin{figure*}[t]
    \centering
    \includegraphics[width=0.8\textwidth]{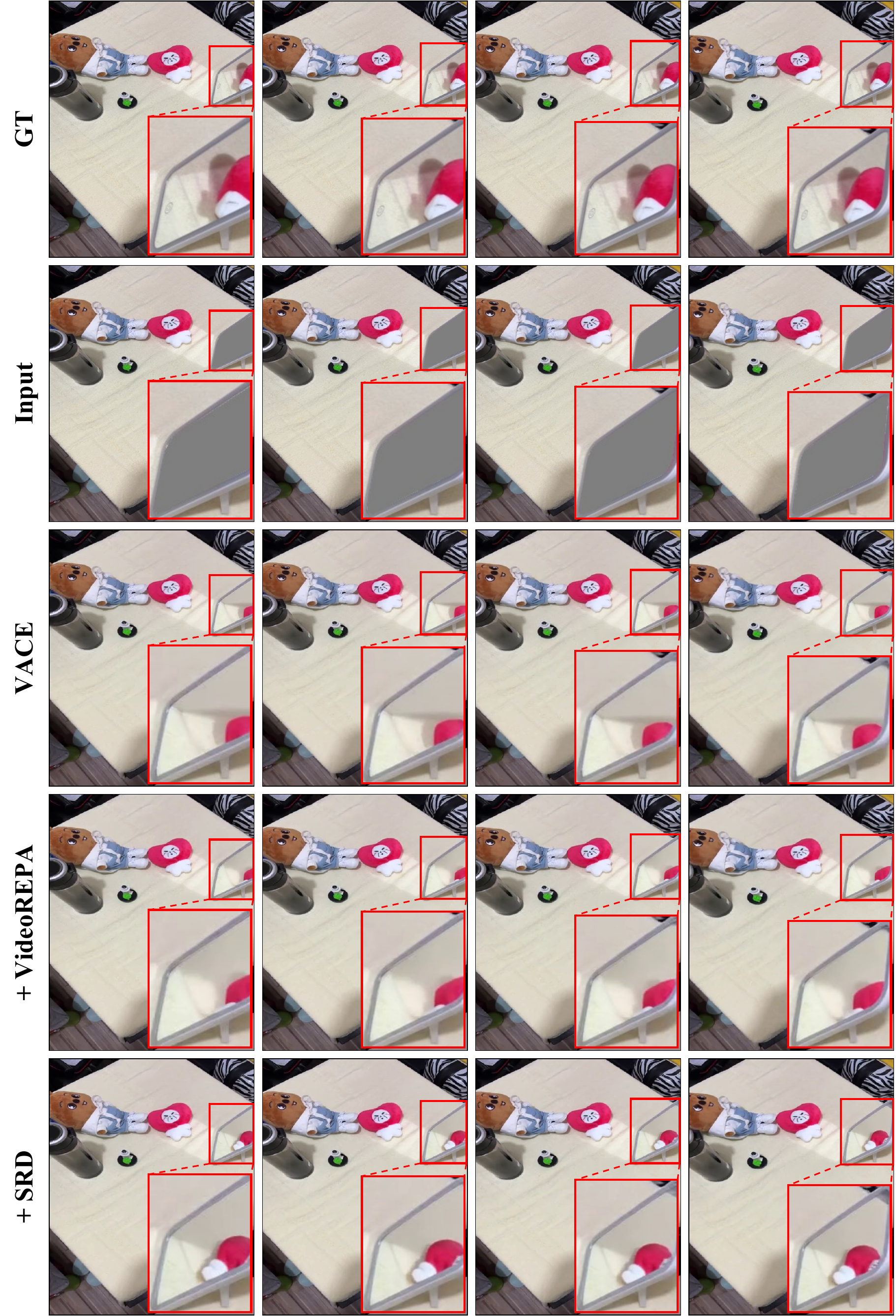}
    \caption{\textbf{Qualitative comparison of relation objectives.} The mirror region is marked by a red box and enlarged in the bottom-right corner of each frame. VACE~\cite{vace} and VACE with VideoREPA~\cite{videorepa} recover only the red cap of the reflected toy, whereas VACE with SRD also reconstructs the white glove beneath it, matching the ground truth.}
    \label{fig:supp_srd_qual}
\end{figure*}

\begin{figure*}[t]
    \centering
    \includegraphics[width=\textwidth]{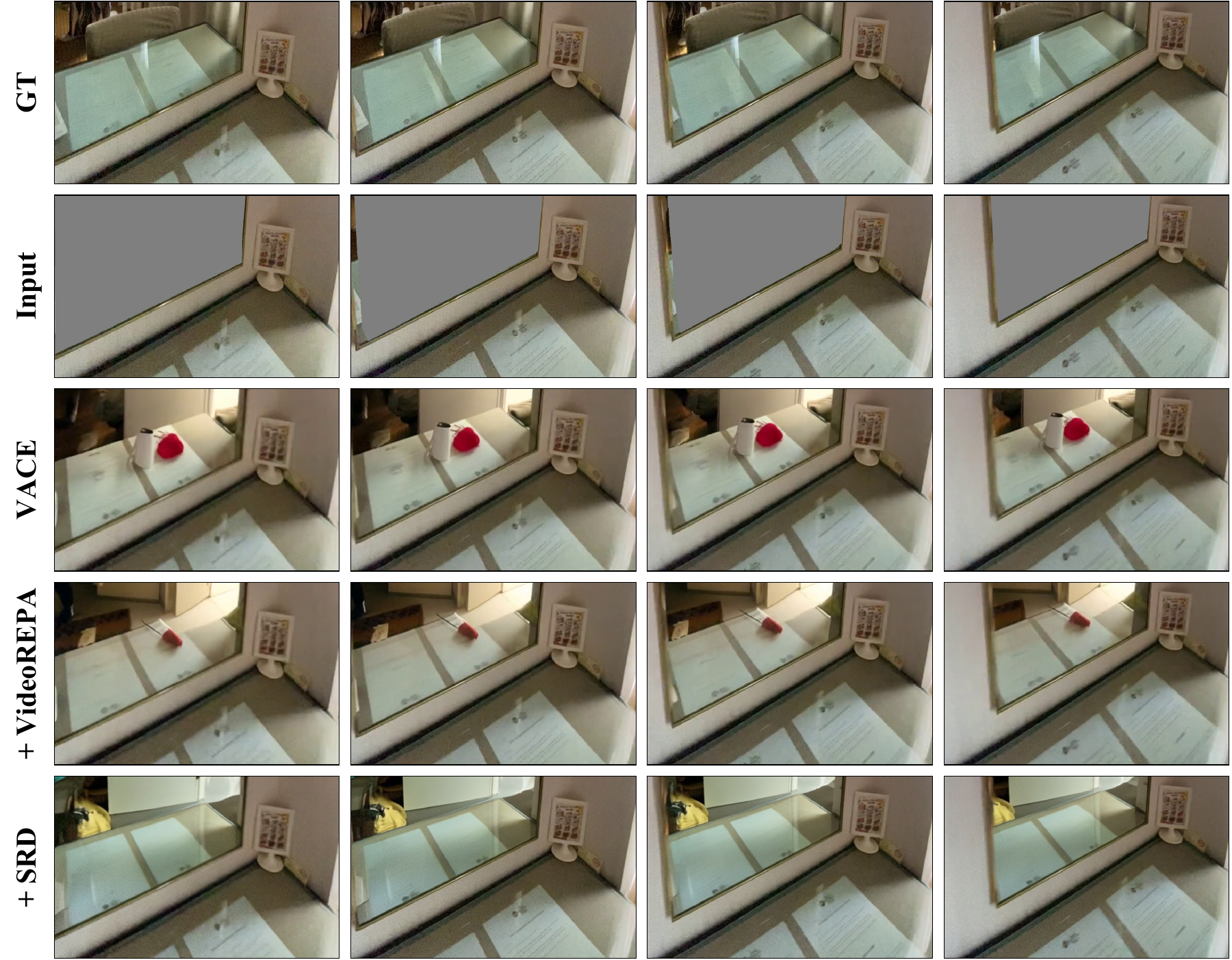}
    \caption{\textbf{Qualitative comparison of relation objectives.}
VACE~\cite{vace} and VACE with VideoREPA~\cite{videorepa} generate extra objects on the two reflected sheets, whereas VACE with SRD reconstructs both sheets cleanly.}
    \label{fig:supp_srd_qual_2}
\end{figure*}

Figure~\ref{fig:supp_srd_qual} compares VACE~\cite{vace} against variants trained with VideoREPA~\cite{videorepa} and SRD. We train all three models without GTA and sample them under identical settings. In this scene, a plush toy rests on a table next to a small mirror, where the ground-truth reflection contains both the toy's red cap and the white glove beneath it. Both VACE and VideoREPA synthesize the red cap but omit the white glove across all displayed frames, whereas SRD reconstructs both parts. This qualitative behavior aligns with the higher $\mathrm{Vis.\,Corr.}$ in Table~\ref{tab:supp_selective}, illustrating more complete recovery of valid reflected objects.
Figure~\ref{fig:supp_srd_qual_2} shows a second clip where two printed sheets on a glass tabletop appear both directly and in the mirror. On the reflected sheets, VACE and VideoREPA generate objects absent from both the visible scene and the ground truth, specifically a white cup with a red object and a red-tipped stick, reflecting the failure mode measured by the hallucination rate $\mathrm{Halluc.}$ in Table~\ref{tab:relation_placement} of the main paper. Conversely, SRD reconstructs both sheets cleanly without such objects, consistent with its lowest hallucination rate.

Thus, the semantic gain of SRD over VideoREPA in Table~\ref{tab:relation_placement} of the main paper is consistent with selective recovery of valid reflected content rather than indiscriminate copying of the visible scene or hallucination of non-existent objects.
Specifically, SRD directly supervises the mirror-to-visible token relations while holding visible tokens as fixed anchors. In contrast, since the mirror occupies only a small part of the frame, most of the token pairs aligned by VideoREPA fall entirely within the visible background and provide no cross-region supervision for reflection synthesis.

\begin{table}[t]
\centering
\small
\begin{tabular}{lccc}
\toprule
& \textbf{Vis.\ Corr.} $\uparrow$ & \textbf{Copy} $\downarrow$ & $\textbf{Copy}^{\ast}$ $\downarrow$ \\
\midrule
VACE~\cite{vace}             & 0.468 & 0.140 & 0.119 \\
+ VideoREPA~\cite{videorepa} & 0.452 & \textbf{0.130} & 0.115 \\
\textbf{+ SRD}               & \textbf{0.488} & 0.131 & \textbf{0.108} \\
\bottomrule
\end{tabular}
\caption{\textbf{Selective recovery of visible content without GTA.} $\mathrm{Vis.\,Corr.}$ denotes the fraction of generated mirror categories present in both the visible scene and ground truth, $\mathrm{Copy}$ denotes those absent from the ground truth, and $\mathrm{Copy}^{\ast}$ restricts $\mathrm{Copy}$ to categories with zero detected pixels in the ground-truth mirror region. Best results are in \textbf{bold}.}
\label{tab:supp_selective}
\end{table}

\subsection{Effect of GTA on Relation Alignment}
\label{sec:supp_videorepa}

SRD and GTA target complementary aspects of reflection reconstruction: the semantic analyses in Table~\ref{tab:relation_placement} of the main paper and Sec.~\ref{sec:supp_selective} are consistent with SRD primarily improving reflected content, while GTA adds a feature-space constraint on spatial arrangement.
To test whether GTA's benefit depends on the paired relation objective, we combine GTA with SRD and with the region-agnostic relation alignment of VideoREPA~\cite{videorepa} under identical settings. Unlike VideoREPA, which aligns relations among all tokens and updates every token, SRD supervises only mirror-to-visible relations with a stop-gradient on the visible tokens.

In Table~\ref{tab:supp_videorepa}, without GTA, VideoREPA and SRD achieve comparable mirror PSNR at 13.521 and 13.551, both close to VACE's 13.537~\cite{vace}, while VideoREPA mainly lowers FVD. Adding GTA improves mirror-region reconstruction for both objectives, raising PSNR from 13.521 to 13.766 with VideoREPA and from 13.551 to 14.006 with SRD alongside better SSIM and LPIPS. This confirms MirrorWorld's decomposition: SRD improves reflected semantics, as analyzed in Table~\ref{tab:relation_placement} of the main paper and Sec.~\ref{sec:supp_selective}, whereas GTA improves pixel-level reconstruction as a module complementary to relation alignment rather than tied to a specific relation loss.

\begin{table}[t]
    \centering
    \resizebox{\linewidth}{!}
    {
    \begin{tabular}{lcccc}
        \toprule
        \textbf{Method}
        & \textbf{PSNR} $\uparrow$
        & \textbf{SSIM} $\uparrow$
        & \textbf{LPIPS} $\downarrow$
        & \textbf{FVD} $\downarrow$ \\
        \midrule
        VACE~\cite{vace}
        & 13.537
        & 0.489
        & 0.493
        & 191.617 \\
        \midrule
        + VideoREPA
        & 13.521
        & 0.494
        & 0.494
        & \textbf{179.281} \\
        + VideoREPA + GTA
        & 13.766
        & 0.500
        & \textbf{0.488}
        & 187.911 \\
        \midrule
        + SRD
        & 13.551
        & 0.500
        & 0.489
        & 185.708 \\
        \textbf{+ SRD + GTA}
        & \textbf{14.006}
        & \textbf{0.504}
        & \textbf{0.488}
        & 184.868 \\
        \bottomrule
    \end{tabular}
    }
    \caption{Effect of GTA on relation alignment. GTA is added to the region-agnostic relation alignment of VideoREPA~\cite{videorepa} and to SRD. PSNR, SSIM, and LPIPS are computed in the mirror region, and FVD is computed on the full frame. Best results are in \textbf{bold}.}
    \label{tab:supp_videorepa}
\end{table}

\section{Additional Ablation Study}
\label{sec:supp_ablation}

\subsection{Backbone Generalization}
\label{sec:supp_backbone}

To evaluate whether our improvements generalize across backbones, we apply the same training recipe to Wan2.2-VACE-Fun-A14B~\cite{wan}, which employs separate high-noise and low-noise expert models. For both fine-tuned VACE and MirrorWorld, each expert is trained for the same iterations within its designated timestep range using identical data and hyperparameters, ensuring that the two models differ solely in the SRD and GTA objectives. As shown in Table~\ref{tab:supp_backbone}, MirrorWorld improves PSNR by 0.44\,dB and SSIM by 0.011 over fine-tuned VACE on Wan2.2, alongside lower LPIPS and FVD. This gain is close to that on Wan2.1 (+0.47\,dB PSNR), supporting transfer to a second tested backbone. We adopt Wan2.1 as our primary backbone as Wan2.2 yields lower baseline performance on this task.

\begin{table}[t]
    \centering
    \resizebox{\linewidth}{!}
    {
    \begin{tabular}{lcccc}
        \toprule
        \textbf{Method}
        & \textbf{PSNR} $\uparrow$
        & \textbf{SSIM} $\uparrow$
        & \textbf{LPIPS} $\downarrow$
        & \textbf{FVD} $\downarrow$ \\
        \midrule
        \rowcolor{gray!10}
        \multicolumn{5}{l}{\textit{\textbf{Wan2.1-VACE}}} \\
        VACE~\cite{vace}
        & 13.537
        & 0.489
        & 0.493
        & 191.6 \\
        \textbf{Ours}
        & \textbf{14.006}
        & \textbf{0.504}
        & \textbf{0.488}
        & \textbf{184.9} \\
        \midrule
        \rowcolor{gray!10}
        \multicolumn{5}{l}{\textit{\textbf{Wan2.2-VACE}}} \\
        VACE~\cite{vace}
        & 12.764
        & 0.485
        & 0.521
        & 207.9 \\
        \textbf{Ours}
        & \textbf{13.199}
        & \textbf{0.496}
        & \textbf{0.520}
        & \textbf{201.9} \\
        \bottomrule
    \end{tabular}
    }
    \caption{Quantitative comparison across two backbones. Best results per backbone are in \textbf{bold}.}
    \label{tab:supp_backbone}
\end{table}

\begin{figure*}[!ht] \centering
    \includegraphics[width=1\textwidth]{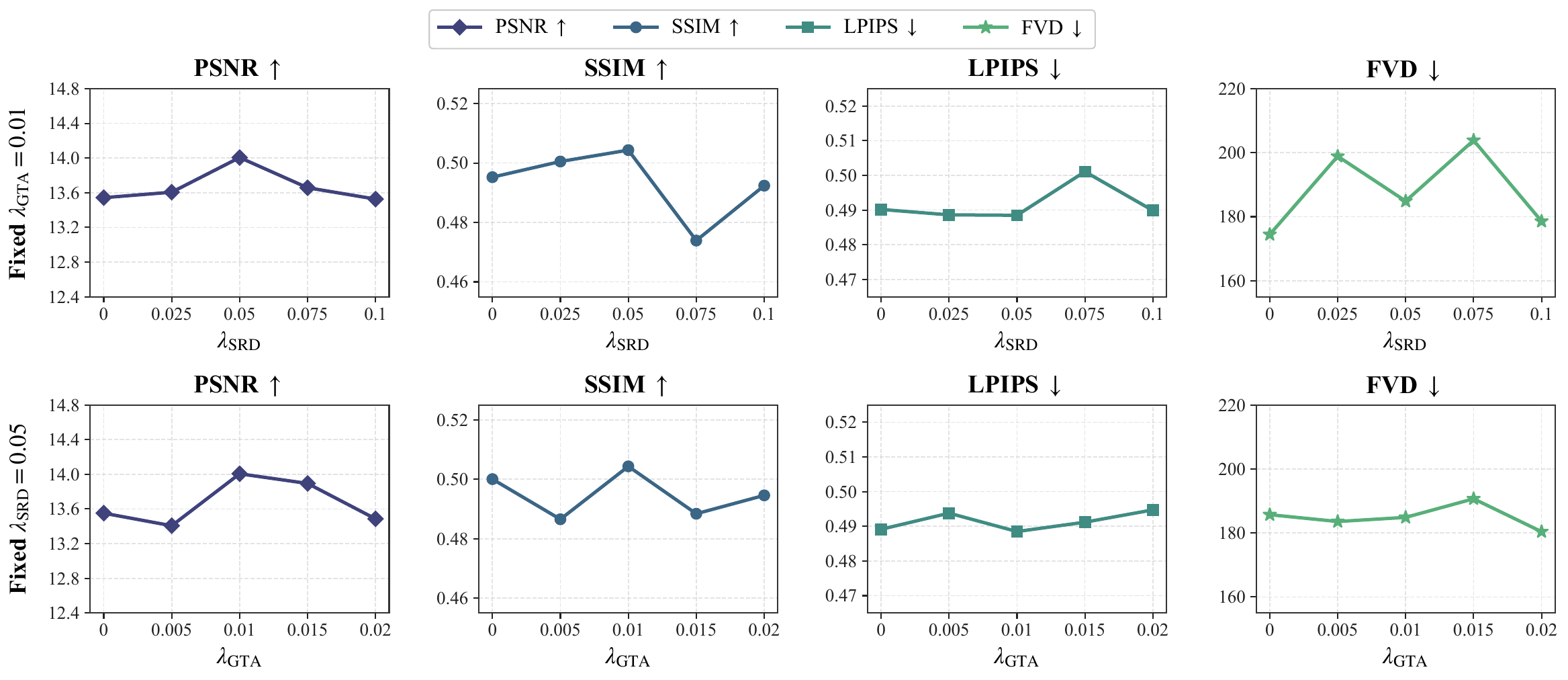}

\caption{\textbf{Ablation on loss weights.}
We vary $\lambda_{\mathrm{SRD}}$ while fixing
$\lambda_{\mathrm{GTA}}=0.01$ (top), and vary $\lambda_{\mathrm{GTA}}$ while
fixing $\lambda_{\mathrm{SRD}}=0.05$ (bottom). PSNR, SSIM, and LPIPS are
evaluated within mirror regions, whereas FVD uses full-frame video features.}
\label{fig:loss_weight_sensitivity}

\end{figure*}

\subsection{Loss Weight}

We study the effects of the loss weights
$\lambda_{\mathrm{SRD}}$ and $\lambda_{\mathrm{GTA}}$ by varying one weight
while keeping the other fixed. As shown in Figure~\ref{fig:loss_weight_sensitivity},
$\lambda_{\mathrm{SRD}}=0.05$ provides the strongest mirror-region
reconstruction across PSNR, SSIM, and LPIPS. Similarly,
$\lambda_{\mathrm{GTA}}=0.01$ performs best on all three reconstruction
metrics. Smaller or larger values lead to less accurate reflected content,
indicating that an intermediate strength is more suitable for both auxiliary
objectives.

The lowest FVD occurs at different weights from the mirror-region metrics.
This difference is expected because FVD evaluates the full-frame video
distribution, whereas PSNR, SSIM, and LPIPS focus on the reconstructed mirror
regions. Therefore, improvements targeted at
reflection correctness may not produce proportional gains in this global
distribution metric. We therefore treat FVD as complementary to the
mirror-region reconstruction metrics.

This distinction also explains the component ablation in
Table~\ref{tab:module_ablation} of the main paper. Although GTA alone obtains the lowest FVD, combining SRD
and GTA achieves the strongest mirror-region reconstruction results. Since FVD
evaluates the distribution of complete frames rather than reflection
correctness specifically, the lowest FVD does not imply that GTA alone is
preferable for video mirror reflection generation.

\begin{table}[t]
    \centering
    \resizebox{\linewidth}{!}
    {
    \begin{tabular}{lcccc}
        \toprule
        \textbf{GTA Token}
        & \textbf{PSNR} $\uparrow$
        & \textbf{SSIM} $\uparrow$
        & \textbf{LPIPS} $\downarrow$
        & \textbf{FVD} $\downarrow$ \\
        \midrule
        Shared with SRD
        & 13.773
        & \textbf{0.504}
        & \textbf{0.484}
        & 196.814 \\
        \textbf{All tokens (Ours)}
        & \textbf{14.006}
        & \textbf{0.504}
        & 0.488
        & \textbf{184.868} \\
        \bottomrule
    \end{tabular}
    }
    \caption{Ablation on token sharing between SRD and GTA.}
    \label{tab:token_sharing}
\end{table}

\subsection{Token Sharing}

We investigate whether SRD and GTA should operate on the same token subset. In
our framework, SRD randomly samples mirror and visible-scene tokens to model
their semantic relations, while GTA uses all tokens to learn the geometric
transformation. The shared variant instead reuses the token locations sampled
by SRD for both components.
As shown in Table~\ref{tab:token_sharing}, sharing tokens slightly improves
LPIPS and preserves SSIM, but reduces PSNR and substantially increases FVD.
This suggests that the sparse token pairs used for semantic relation
distillation do not provide sufficient spatial coverage for learning the
geometric transformation. Using all tokens allows GTA to better capture how
visible content should be spatially arranged, while SRD focuses on learning
what content should be reflected. The LPIPS improvement is relatively small, suggesting that sharing tokens may preserve certain local perceptual properties. However, the reduced PSNR and
substantially worse FVD indicate that the sparse shared token support is less
suitable for geometric alignment and global video-level quality.

\begin{table}[t]
    \centering
    \resizebox{1\linewidth}{!}{
    \begin{tabular}{lcccc}
        \toprule
        \textbf{SRD Token Sampling}
        & \textbf{PSNR} $\uparrow$
        & \textbf{SSIM} $\uparrow$
        & \textbf{LPIPS} $\downarrow$
        & \textbf{FVD} $\downarrow$ \\
        \midrule
        Top-$N$
        & 13.591
        & 0.494
        & 0.491
        & \textbf{173.294} \\
        \textbf{Random (Ours)}
        & \textbf{14.006}
        & \textbf{0.504}
        & \textbf{0.488}
        & 184.868 \\
        \bottomrule
    \end{tabular}
    }
    \caption{Ablation on token sampling strategy in SRD.}
    \label{tab:srd_token_selection}
\end{table}

\begin{table*}[t]
    \centering
    \resizebox{0.7\linewidth}{!}
    {
    \begin{tabular}{lcccccc}
        \toprule
        & \multicolumn{3}{c}{\textbf{VACE~\cite{vace}}} & \multicolumn{3}{c}{\textbf{Ours}} \\
        \cmidrule(lr){2-4} \cmidrule(lr){5-7}
        \textbf{Source Dataset}
        & PSNR $\uparrow$ & SSIM $\uparrow$ & LPIPS $\downarrow$
        & PSNR $\uparrow$ & SSIM $\uparrow$ & LPIPS $\downarrow$ \\
        \midrule
        DVMD-D~\cite{DVMD-D} & 14.756 & 0.537 & 0.462 & \textbf{15.295} & \textbf{0.551} & \textbf{0.453} \\
        ZOOM~\cite{zoom}   & 10.555 & 0.357 & \textbf{0.537} & \textbf{11.024} & \textbf{0.385} & 0.542 \\
        MMD~\cite{mmd}    & 13.146 & \textbf{0.510} & \textbf{0.511} & \textbf{13.289} & 0.509 & 0.512 \\
        VMD-D~\cite{vmd}  & 11.762 & 0.404 & 0.594 & \textbf{12.189} & \textbf{0.419} & \textbf{0.591} \\
        \midrule
        All    & 13.537 & 0.489 & 0.493 & \textbf{14.006} & \textbf{0.504} & \textbf{0.488} \\
        \bottomrule
    \end{tabular}
    }
    \caption{Breakdown of mirror-region quantitative results across the four source datasets in the test set. Best results are highlighted in \textbf{bold}.}
    \label{tab:supp_per_dataset}
\end{table*}

\subsection{Token Sampling}

We investigate the token sampling strategy used to construct the scene-to-mirror relations in SRD. For random sampling, we independently sample token indices from both the mirror region $\mathcal{I}_{\tau}$ and the visible region $\mathcal{O}_{\tau}$. The same sampled index pairs are used to construct the reference and diffusion relation matrices $\mathbf{R}^{R}_{\tau}$ and $\mathbf{R}^{D}_{\tau}$, respectively.

For Top-$N$ sampling, the mirror indices are sampled in the same manner. For each sampled mirror token $i\in\mathcal{I}_{\tau}$, we compute its reference relations with all visible-region tokens and select the $N$ indices with the highest relation scores $\mathbf{R}^{R}_{\tau}(i,j)$. The selected index pairs are then shared with the diffusion branch to compute the corresponding relations $\mathbf{R}^{D}_{\tau}(i,j)$. Thus, the two strategies differ only in how the visible-region tokens are selected, while all other settings remain unchanged. In our experiments, random sampling uses up to 64 tokens from each region, while Top-$N$ sampling uses $N=16$.

As shown in Table~\ref{tab:srd_token_selection}, random sampling performs better across all mirror-region reconstruction metrics. This suggests that covering a broader range of scene-to-mirror relations is more effective for reconstructing complex reflections than focusing only on the strongest reference relations. Although Top-$N$ sampling achieves a lower FVD, this improvement in video-level distribution quality does not benefit the reconstruction within the mirror regions.

\section{Additional Experiments and Analyses}
\label{sec:supp_experiments}

\subsection{Evaluation Across Source Datasets}
\label{sec:supp_per_dataset}
Our test set comprises clips drawn from four source datasets. Table~\ref{tab:supp_per_dataset} breaks down the mirror-region metrics of VACE~\cite{vace} and our MirrorWorld across each source. MirrorWorld improves mirror-region PSNR across all four subsets, confirming that the overall performance gain is not driven by any single dataset.

\subsection{Comparison with a Training-Free Geometric Baseline}
\label{sec:supp_geometric}

To examine whether a fixed geometric heuristic suffices to synthesize mirror reflections, we construct a training-free flip-and-paste baseline that horizontally reflects visible pixels across the vertical axis passing through the mirror center and pastes them into the mirror region, with unmapped mirror pixels filled via Telea inpainting. This baseline relies solely on visible pixels, matching the input available to our model. As shown in Table~\ref{tab:supp_geometric}, flip-and-paste performs substantially worse than our model across all metrics, particularly in PSNR, LPIPS, and FVD. Most reflected coordinates fall outside the frame or within the mirror mask, leaving only 16\% of mirror pixels with valid visible content on average, and even the copied pixels rarely align with the true 3D viewpoint. Mirror reflection synthesis therefore cannot be reduced to a fixed 2D image-space transformation of visible pixels.

\begin{table}[h]
    \centering
    \resizebox{\linewidth}{!}
    {
    \begin{tabular}{lcccc}
        \toprule
        \textbf{Method}
        & \textbf{PSNR} $\uparrow$
        & \textbf{SSIM} $\uparrow$
        & \textbf{LPIPS} $\downarrow$
        & \textbf{FVD} $\downarrow$ \\
        \midrule
        Flip-and-paste
        & 11.725
        & 0.500
        & 0.612
        & 303.2 \\
        \textbf{Ours}
        & \textbf{14.006}
        & \textbf{0.504}
        & \textbf{0.488}
        & \textbf{184.9} \\
        \bottomrule
    \end{tabular}
    }
    \caption{Quantitative comparison with a training-free geometric baseline that relies solely on visible pixels. Best results are highlighted in \textbf{bold}.}
    \label{tab:supp_geometric}
\end{table}

\begin{figure*}[!t] \centering
    \includegraphics[width=1\textwidth]{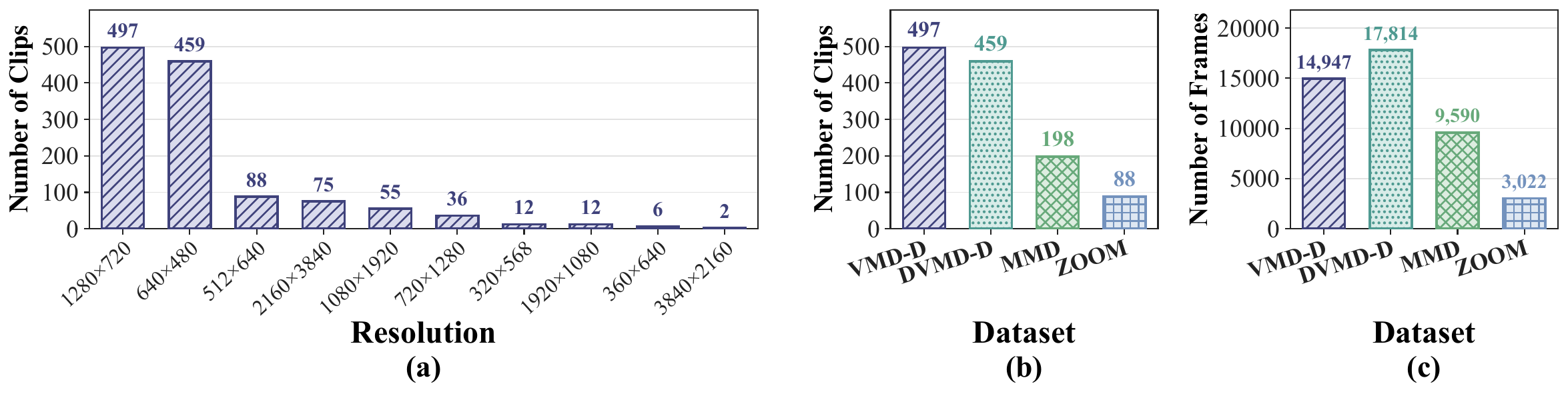}
\caption{\textbf{Statistics of our benchmark.}
(a) Distribution of clip resolutions. (b) Number of clips and
(c) number of frames contributed by each dataset.}
\label{fig:benchmark_statistics}
\end{figure*}

\section{Benchmark Statistics}
\label{sec:benchmark_statistics}

Our benchmark integrates four video mirror detection and segmentation datasets, including VMD-D~\cite{vmd}, DVMD-D~\cite{DVMD-D}, MMD~\cite{mmd}, and ZOOM~\cite{zoom}, resulting in $1{,}242$ video clips comprising $45{,}373$ frames. As shown in Figure~\ref{fig:benchmark_statistics}, these clips cover a wide range of spatial resolutions. By using these datasets under a unified reflection reconstruction setting, our benchmark provides a larger and more diverse data resource for training and systematically evaluating video mirror reflection generation methods.

In our benchmark, split assignments are determined using the identities of the original source
videos. All clips derived from the same source video are assigned exclusively
to either the training or testing partition, and no source video contributes
clips to both partitions. This source-level partitioning prevents content
leakage caused by placing different clips from the same video in separate
splits.

Since the source datasets contain real videos with visible mirrors, the
original unmasked clips directly provide the ground-truth reflections. During
benchmark construction, mirror regions are masked only in the conditional
inputs, while the original clips are retained as the supervision and
evaluation targets.

\section{Formulation of MirrorWorld}
\label{sec:supp_formulation}

\subsection{Preliminaries}
\label{sec:supp_prelim}

We briefly review the flow-matching formulation and conditioning mechanism of our Wan2.1-VACE backbone~\cite{wan} and define the notation used in Algorithms~\ref{alg:mirrorworld_train} and~\ref{alg:mirrorworld_infer}.

\noindent\textbf{Flow matching.}
A video $\mathbf{V}$ is encoded into a latent $\mathbf{z}_{0}=\mathcal{E}_{\mathrm{vae}}(\mathbf{V})$ by a video VAE with decoder $\mathcal{D}_{\mathrm{vae}}$. Flow matching~\cite{lipman2023flow,liu2023flow} interpolates between the data latent and Gaussian noise $\boldsymbol{\epsilon}\sim\mathcal{N}(\mathbf{0},\mathbf{I})$ along a straight path. At diffusion timestep $s$ with noise level $\sigma_{s}\in[0,1]$,
\begin{equation}
    \mathbf{z}_{s}
    =
    (1-\sigma_{s})\,\mathbf{z}_{0}
    +
    \sigma_{s}\,\boldsymbol{\epsilon},\label{eq:supp_forward}
\end{equation}
which instantiates $q_{s}$ in \mainsec{Sec.~\ref{sec:method}}. The network $\mathbf{u}_{\theta}$ is trained to predict the path velocity,
\begin{equation}
    \mathbf{u}^{*}_{s}
    =
    \frac{\mathrm{d}\mathbf{z}_{s}}{\mathrm{d}\sigma_{s}}
    =
    \boldsymbol{\epsilon}-\mathbf{z}_{0},\label{eq:supp_target}
\end{equation}
which serves as the target in the diffusion loss of \mainsec{Eq.~\eqref{eq:diffusion_loss}}.

\noindent\textbf{Sampling.}
Generation starts from pure noise $\mathbf{z}_{s_{0}}\sim\mathcal{N}(\mathbf{0},\mathbf{I})$ and integrates $\mathrm{d}\mathbf{z}/\mathrm{d}\sigma=\mathbf{u}_{\theta}$ from $\sigma=1$ to $\sigma=0$ via an Euler solver over a decreasing schedule $1=\sigma_{0}>\sigma_{1}>\dots>\sigma_{J}=0$, where $s_{j}$ denotes the timestep corresponding to $\sigma_{j}$. Each step applies classifier-free guidance~\cite{ho2022classifier}:
\begin{align}
    \tilde{\mathbf{u}}_{j}
    &=
    \mathbf{u}^{-}_{j}
    +
    \omega\,\big(\mathbf{u}^{+}_{j}-\mathbf{u}^{-}_{j}\big),\label{eq:supp_cfg}
    \\
    \mathbf{z}_{s_{j+1}}
    &=
    \mathbf{z}_{s_{j}}
    +
    (\sigma_{j+1}-\sigma_{j})\,\tilde{\mathbf{u}}_{j},\label{eq:supp_euler}
\end{align}
where $\mathbf{u}^{+}_{j}=\mathbf{u}_{\theta}(\mathbf{z}_{s_{j}},s_{j},\mathbf{c})$, $\mathbf{u}^{-}_{j}=\mathbf{u}_{\theta}(\mathbf{z}_{s_{j}},s_{j},\mathbf{c}^{-})$, and $\omega$ is the guidance scale. The negative condition $\mathbf{c}^{-}$ replaces the text prompt $\mathbf{T}$ with a negative prompt. The synthesized video is obtained as $\widehat{\mathbf{V}}=\mathcal{D}_{\mathrm{vae}}(\mathbf{z}_{s_{J}})$.

\noindent\textbf{Conditioning.}
VACE~\cite{vace} introduces a context branch into the DiT for video editing tasks such as inpainting. Given the condition $\mathbf{c}=(\mathbf{V}^{m},\mathbf{M},\mathbf{T})$, it encodes the preserved visible region, the masked target region, and the mask into a context latent:
\begin{equation}
    \mathbf{z}^{c}
    =
    \Big[\,
    \mathcal{E}_{\mathrm{vae}}\big(\mathbf{V}^{m}\odot(1-\mathbf{M})\big);\,
    \mathcal{E}_{\mathrm{vae}}\big(\mathbf{V}^{m}\odot\mathbf{M}\big);\,
    \mathcal{P}(\mathbf{M})
    \,\Big],\label{eq:supp_vace_context}
\end{equation}
where $[\,\cdot\,;\,\cdot\,]$ denotes channel-wise concatenation and $\mathcal{P}$ reshapes the mask to the latent resolution. The VACE blocks process $\mathbf{z}^{c}$ and inject their output hints into the main DiT hidden states, while the text prompt $\mathbf{T}$ is incorporated via cross-attention. MirrorWorld updates only the LoRA weights within the VACE blocks while keeping the remaining backbone parameters frozen.

\subsection{Formulation of SRD}
\label{sec:supp_srd_formulation}

Let $\bar{\mathbf{F}}^{D}$ and $\bar{\mathbf{F}}^{R}$ denote the $\ell_2$-normalized student and teacher features on the teacher token grid, so that $\mathbf{G}=\bar{\mathbf{F}}\bar{\mathbf{F}}^{\top}$ is their cosine relation matrix. Splitting the tokens into the mirror set $\mathcal{I}$ and the visible set $\mathcal{O}$ partitions $\mathbf{G}$ into four blocks:
\begin{equation}
\mathbf{G}=\begin{bmatrix}\mathbf{G}_{\mathcal{I}\mathcal{I}} & \mathbf{G}_{\mathcal{I}\mathcal{O}}\\ \mathbf{G}_{\mathcal{O}\mathcal{I}} & \mathbf{G}_{\mathcal{O}\mathcal{O}}\end{bmatrix}.
\label{eq:supp_gram_blocks}
\end{equation}
Our adaptation of VideoREPA~\cite{videorepa} aligns the entire $N\times N$ matrix over all $N=T'hw$ space-time tokens with a margin $\delta=0.1$, and both factors of $\mathbf{G}^{D}$ receive gradients:
\begin{equation}
\begin{aligned}
\mathcal{L}_{\mathrm{VR}}&=\frac{1}{N^{2}}\Big\|\max\big(0,\,\big|\mathbf{G}^{D}-\mathbf{G}^{R}\big|-\delta\big)\Big\|_{1},\\
\mathbf{G}^{D}&=\bar{\mathbf{F}}^{D}\,\bar{\mathbf{F}}^{D\top}.
\end{aligned}
\label{eq:supp_vr}
\end{equation}
SRD (\mainsec{Eq.~\eqref{eq:srd}}) keeps only the mirror--visible block within each temporal slice $\tau$ and treats the visible factor as a constant:
\begin{equation}
\begin{aligned}
\mathcal{L}_{\mathrm{SRD}}&=\frac{1}{|\mathcal{V}_{S}|}\sum_{\tau\in\mathcal{V}_{S}}\frac{\big\|\mathbf{G}^{D}_{\tau,\mathcal{I}\mathcal{O}}-\mathbf{G}^{R}_{\tau,\mathcal{I}\mathcal{O}}\big\|_{1}}{|\mathcal{I}_{\tau}|\,|\mathcal{O}_{\tau}|},\\
\mathbf{G}^{D}_{\tau,\mathcal{I}\mathcal{O}}&=\bar{\mathbf{F}}^{D}_{\tau,\mathcal{I}}\,\operatorname{sg}\big[\bar{\mathbf{F}}^{D}_{\tau,\mathcal{O}}\big]^{\top}.
\end{aligned}
\label{eq:supp_srd}
\end{equation}
The two equations differ in four aspects. (1) \emph{Supervised entries.} $\mathcal{L}_{\mathrm{VR}}$ is dominated by $\mathbf{G}_{\mathcal{O}\mathcal{O}}$: when the mirror covers a fraction $\alpha$ of the tokens, the visible--visible block contributes $(1-\alpha)^{2}$ of its terms and the mirror--visible blocks only $2\alpha(1-\alpha)$, which is about 65\% and 29\% averaged over our test clips. Since the visible content is given as input, most of $\mathcal{L}_{\mathrm{VR}}$ supervises relations that carry little information about the reflection. $\mathcal{L}_{\mathrm{SRD}}$ supervises only $\mathbf{G}_{\mathcal{I}\mathcal{O}}$, which encodes how the reflection relates to the scene. (2) \emph{Gradient.} In $\mathcal{L}_{\mathrm{VR}}$, every token appears in both factors of $\mathbf{G}^{D}$ and is updated through all four blocks. In $\mathcal{L}_{\mathrm{SRD}}$, $\partial\mathcal{L}_{\mathrm{SRD}}/\partial\bar{\mathbf{F}}^{D}_{\tau,j}=\mathbf{0}$ for every visible token $j\in\mathcal{O}_{\tau}$, so only the mirror features are moved toward the teacher relations. (3) \emph{Temporal support.} $\mathcal{L}_{\mathrm{VR}}$ relates tokens across all frames, whereas $\mathcal{L}_{\mathrm{SRD}}$ relates mirror and visible tokens of the same temporal slice. (4) \emph{Error.} The margin in $\mathcal{L}_{\mathrm{VR}}$ ignores relation errors below $\delta$, whereas $\mathcal{L}_{\mathrm{SRD}}$ penalizes all of them.

\subsection{Formulation of GTA}
\label{sec:supp_gta_formulation}

Let $\mathbf{F}^{G}_{t}$ denote the GTA features at step $t$, $\mathbf{M}_{t}$ the binary mirror mask, and $\mathcal{W}(\cdot,\mathbf{A})$ differentiable bilinear warping with affine matrix $\mathbf{A}$. Following \mainsec{Eq.~\eqref{eq:gta_source}}, the source map $\mathbf{S}_{t}$ retains visible features in $\mathcal{O}_{t}$ and fills mirror locations with their mean $\boldsymbol{\mu}_{t}$:
\begin{equation}
\begin{aligned}
\mathbf{S}_{t}&=(1-\mathbf{M}_{t})\odot\mathbf{F}^{G}_{t}+\mathbf{M}_{t}\odot\boldsymbol{\mu}_{t},\\
\boldsymbol{\mu}_{t}&=\frac{1}{|\mathcal{O}_{t}|}\sum_{\mathbf{p}\in\mathcal{O}_{t}}\mathbf{F}^{G}_{t}(\mathbf{p}).
\end{aligned}
\label{eq:supp_gta_source}
\end{equation}
Inserted into the forward pass, a spatial transformer network (STN)~\cite{stn} employs a localization network $f_{\mathrm{loc}}$ to regress transformation parameters from the input map $\mathbf{U}$ and a differentiable sampler $\mathcal{W}$ to resample $\mathbf{U}$ before passing it to the next layer $h$, supervised solely by the task loss:
\begin{equation}
\mathcal{L}_{\mathrm{STN}}=\mathcal{L}_{\mathrm{task}}\Big(h\big(\mathcal{W}(\mathbf{U},\,f_{\mathrm{loc}}(\mathbf{U}))\big)\Big).
\label{eq:supp_stn}
\end{equation}
In contrast, GTA places the warp inside the training loss in \mainsec{Eq.~\eqref{eq:gta_loss}} and aligns the warped source map with the mirror features:
\begin{equation}
\begin{aligned}
\mathcal{L}_{\mathrm{GTA}}&=\frac{1}{|\mathcal{V}_{G}|}\sum_{t\in\mathcal{V}_{G}}\frac{1}{|\mathcal{I}_{t}|}\sum_{\mathbf{q}\in\mathcal{I}_{t}}\Big[1-\operatorname{sim}\big(\widehat{\mathbf{F}}^{G}_{t}(\mathbf{q}),\,\mathbf{F}^{G}_{t}(\mathbf{q})\big)\Big],\\
\widehat{\mathbf{F}}^{G}_{t}&=\mathcal{W}\big(\operatorname{sg}[\mathbf{S}_{t}],\,\mathbf{A}_{t}\big),\\
\mathbf{A}_{t}&=g_{\eta}\Big(\frac{1}{K}\sum_{k=-r}^{r}\operatorname{sg}\big[\mathbf{S}_{\operatorname{clip}(t+k)}\big]\Big),
\end{aligned}
\label{eq:supp_gta}
\end{equation}
where $\mathcal{V}_{G}$ denotes supervised steps, $\mathcal{I}_{t}$ is the mirror region, $\operatorname{sim}(\cdot,\cdot)$ is cosine similarity, and $\operatorname{sg}[\cdot]$ is the stop-gradient operator.

The two formulations differ in four aspects:
(1) \emph{Role of the warp.} $\mathcal{L}_{\mathrm{STN}}$ executes warping in the forward pass at inference, whereas $\mathcal{L}_{\mathrm{GTA}}$ applies $\mathcal{W}$ only during training and discards $g_{\eta}$ and $\phi_{\mathrm{GTA}}$ afterward, keeping inference unchanged.
(2) \emph{Input of the warp.} An STN warps its full input, while GTA warps $\mathbf{S}_{t}$ filled with $\boldsymbol{\mu}_{t}$ inside the mirror so that an identity warp yields the visible mean rather than existing mirror features.
(3) \emph{Supervision and gradient.} Unlike an STN trained solely by $\mathcal{L}_{\mathrm{task}}$, $\mathcal{L}_{\mathrm{GTA}}$ supervises mirror locations $\mathbf{q}\in\mathcal{I}_{t}$ against $\mathbf{F}^{G}_{t}$ with stop-gradient source features, updating only $g_{\eta}$ and mirror tokens toward the transported visible features.
(4) \emph{Temporal support.} An STN predicts parameters from the current input alone, whereas $\mathbf{A}_{t}$ aggregates a $K$-step source window and warps only $\mathbf{S}_{t}$.

\section{Additional Implementation Details}
\label{sec:supp_impl}

\subsection{Text Prompt}

To ensure a consistent evaluation setting, we use the same task-level text prompt for all video clips and baseline methods. As shown in Figure~\ref{fig:prompt_template}, the prompt describes the desired properties of the reconstructed reflection. We avoid object-specific descriptions so that the generated mirror content must be inferred from the visible scene and mirror mask rather than explicit semantic cues in the prompt. This unified prompt also prevents differences in prompt design from affecting the comparison between methods. The same prompt is used during training and inference. Since the prompt contains only task-level instructions and no scene-specific object identities, the reconstruction is driven primarily by the visual conditions rather than scene-specific prompt engineering.

\begin{figure}[t]
\centering
\begin{tcolorbox}[
    enhanced, width=\linewidth, arc=0pt, frame hidden, colback=white,
    boxsep=0.5mm, left=1mm, right=1mm, top=0.6mm, bottom=1mm,
    interior code={\fill[pattern=crosshatch, pattern color=promptband!30] (interior.south west) rectangle (interior.north east);},
    overlay={\draw[black!80, line width=0.8pt, sketch] (frame.south west) rectangle (frame.north east);}
]
\centering{\ttfamily\bfseries\large Text Prompt}\par\vspace{-1.6mm}
\begin{tcolorbox}[
    enhanced, arc=0pt, frame hidden, colback=gray!6,
    boxsep=0.5mm, left=1.5mm, right=1.5mm, top=1.5mm, bottom=1.5mm, fontupper=\ttfamily\footnotesize,
    overlay={\draw[black!80, line width=0.6pt, sketch] (frame.south west) rectangle (frame.north east);}
]
\raggedright
\textbf{prompt:} '{}'{}'The mirror region in the scene is naturally completed, with reflections consistent with the surroundings, correct structure, realistic details, and temporally coherent motion.'{}'{}'
\end{tcolorbox}
\end{tcolorbox}
\caption{Text prompt for video mirror reflection generation.}
\label{fig:prompt_template}
\end{figure}

\subsection{Training and Inference Algorithms}\label{sec:supp_algorithms}

Algorithms~\ref{alg:mirrorworld_train} and~\ref{alg:mirrorworld_infer} summarize the training and inference procedures, where $\operatorname{sg}[\cdot]$ denotes the stop-gradient operator. Both SRD and GTA reuse the intermediate features $\mathbf{H}_{\theta}$ extracted during the single forward pass used to compute $\mathcal{L}_{\mathrm{diff}}$. By detaching visible-region features as fixed anchors, both objectives backpropagate exclusively through mirror-region features. Moreover, since the teacher $\mathcal{E}$, projection heads $\phi_{\mathrm{SRD}}$ and $\phi_{\mathrm{GTA}}$, and regressor $g_{\eta}$ are active solely during training, the inference pipeline and computational cost remain identical to those of the base VACE backbone.

\begin{algorithm*}[t]
\caption{Training of MirrorWorld.}
\label{alg:mirrorworld_train}
\begin{algorithmic}[1]
\Require Training clip $(\mathbf{V},\mathbf{V}^{m},\mathbf{M},\mathbf{T})$; frozen VAE $\mathcal{E}_{\mathrm{vae}}$, main DiT, and teacher $\mathcal{E}$; trainable LoRA weights $\theta_{\mathrm{L}}$, heads $\phi_{\mathrm{SRD}}$ and $\phi_{\mathrm{GTA}}$, and regressor $g_{\eta}$
\Statex \textit{Diffusion loss}
\State $\mathbf{z}_{0}\gets\mathcal{E}_{\mathrm{vae}}(\mathbf{V})$; \ $\mathbf{c}\gets(\mathbf{V}^{m},\mathbf{M},\mathbf{T})$
\State Sample timestep $s$ and $\boldsymbol{\epsilon}\sim\mathcal{N}(\mathbf{0},\mathbf{I})$; \ $\mathbf{z}_{s}\gets(1-\sigma_{s})\,\mathbf{z}_{0}+\sigma_{s}\,\boldsymbol{\epsilon}$ \Comment{Eq.~\eqref{eq:supp_forward}}
\State $(\mathbf{u}_{\theta},\mathbf{H}_{\theta})\gets f_{\theta}(\mathbf{z}_{s},s,\mathbf{c})$ \Comment{$\mathbf{H}_{\theta}$: intermediate DiT features}
\State $\mathcal{L}_{\mathrm{diff}}\gets w(s)\,\operatorname{MSE}(\mathbf{u}_{\theta},\,\boldsymbol{\epsilon}-\mathbf{z}_{0})$ \Comment{Eq.~\eqref{eq:supp_target}}
\Statex \textit{Semantic Relation Distillation (SRD)}
\State $\mathbf{F}^{R}\gets\mathcal{E}(\mathbf{V})$; \ $\mathbf{F}^{D}\gets\phi_{\mathrm{SRD}}(\mathbf{H}_{\theta})$, resampled onto the token grid of $\mathbf{F}^{R}$
\For{each temporal slice $\tau\in\mathcal{V}_{S}$}
    \State Sample mirror tokens $\mathcal{I}_{\tau}$ and visible tokens $\mathcal{O}_{\tau}$
    \State $\mathbf{R}^{R}_{\tau}(i,j)\gets\operatorname{sim}(\mathbf{F}^{R}_{\tau,i},\mathbf{F}^{R}_{\tau,j})$, \ $\mathbf{R}^{D}_{\tau}(i,j)\gets\operatorname{sim}\big(\mathbf{F}^{D}_{\tau,i},\operatorname{sg}[\mathbf{F}^{D}_{\tau,j}]\big)$ \ for $i\in\mathcal{I}_{\tau}$, $j\in\mathcal{O}_{\tau}$
\EndFor
\State Compute $\mathcal{L}_{\mathrm{SRD}}$ \Comment{\mainsec{Eq.~\eqref{eq:srd}}}
\Statex \textit{Geometric Transformation Alignment (GTA)}
\State $\mathbf{F}^{G}\gets\phi_{\mathrm{GTA}}(\mathbf{H}_{\theta})$
\State $\mathbf{S}_{t}\gets\operatorname{sg}\big[\,\text{\mainsec{Eq.~\eqref{eq:gta_source}}}\,\big]$ for each temporal index $t$ \Comment{mirror positions mean-filled}
\For{each temporal index $t$}
    \State $\mathbf{C}_{t}\gets\frac{1}{K}\sum_{k=-r}^{r}\mathbf{S}_{\operatorname{clip}(t+k)}$; \ $\mathbf{A}_{t}\gets g_{\eta}(\mathbf{C}_{t})$; \ $\widehat{\mathbf{F}}^{G}_{t}\gets\mathcal{W}(\mathbf{S}_{t},\mathbf{A}_{t})$
\EndFor
\State Compute $\mathcal{L}_{\mathrm{GTA}}$ between $\widehat{\mathbf{F}}^{G}_{t}$ and $\mathbf{F}^{G}_{t}$ at mirror positions \Comment{\mainsec{Eq.~\eqref{eq:gta_loss}}}
\Statex \textit{Parameter update}
\State Compute total loss $\mathcal{L}$ \Comment{\mainsec{Eq.~\eqref{eq:overall_objective}}}
\State Update $\theta_{\mathrm{L}}$, $\phi_{\mathrm{SRD}}$, $\phi_{\mathrm{GTA}}$, and $\eta$ via a gradient step on $\mathcal{L}$
\end{algorithmic}
\end{algorithm*}

\begin{algorithm}[t]
\caption{Inference of MirrorWorld. Since $\mathcal{E}$, $\phi_{\mathrm{SRD}}$, $\phi_{\mathrm{GTA}}$, and $g_{\eta}$ are omitted at test time, inference incurs zero auxiliary overhead.}
\label{alg:mirrorworld_infer}
\begin{algorithmic}[1]
\Require $\mathbf{V}^{m}$, $\mathbf{M}$, $\mathbf{T}$; trained LoRA weights $\theta_{\mathrm{L}}$; schedule $\{\sigma_{j}\}_{j=0}^{J}$; guidance scale $\omega$
\State $\mathbf{c}\gets(\mathbf{V}^{m},\mathbf{M},\mathbf{T})$; \ $\mathbf{z}\sim\mathcal{N}(\mathbf{0},\mathbf{I})$
\For{$j=0,\dots,J-1$}
    \State $\tilde{\mathbf{u}}_{j}\gets$ Eq.~\eqref{eq:supp_cfg}
    \State $\mathbf{z}\gets\mathbf{z}+(\sigma_{j+1}-\sigma_{j})\,\tilde{\mathbf{u}}_{j}$ \Comment{Eq.~\eqref{eq:supp_euler}}
\EndFor
\State \Return $\widehat{\mathbf{V}}=\mathcal{D}_{\mathrm{vae}}(\mathbf{z})$ \Comment{direct decoding without pixel copy-pasting}
\end{algorithmic}
\end{algorithm}

\subsection{Training and Inference Details of MirrorWorld}

During training, we preserve the original aspect ratio of each video while
limiting the spatial resolution of each frame to at most $399{,}360$ pixels.
We adapt the backbone using rank-32 LoRA, applied to the query, key, value,
output, and feed-forward projections of the transformer. All pretrained
backbone parameters remain frozen throughout training. SRD and GTA operate on
the same diffusion features, which are separately projected to 256 dimensions
for their respective objectives. At each temporal step, we randomly sample up
to 64 tokens from the mirror and visible regions for SRD.

During inference, the auxiliary SRD and GTA branches are removed, introducing
no additional inference overhead. We generate videos of up to 49 frames using
50 denoising steps and a classifier-free guidance scale of 5.0. The original
aspect ratio is preserved, with the spatial resolution limited to at most
$399{,}360$ pixels per frame. No post-processing or explicit copy-and-paste
operation is applied to the visible regions. Inference takes about 6 minutes per video on a single A100 GPU, the same as the VACE baseline.

\subsection{Baseline Details}

All baselines are fine-tuned using their officially released implementations. Original architectures and training settings are preserved wherever possible, with only minimal modifications required to support the unified benchmark protocol.
All methods are fine-tuned for the same number of epochs as our method on the same training split and evaluated using the same masked videos, mirror masks, text prompts, and testing protocol.

\subsubsection{MirrorFusion and MirrorVerse}

We initialize MirrorFusion~\cite{mirrorfusion} and
MirrorVerse~\cite{mirrorverse} from the released v1 and v2 checkpoints,
respectively. Each video frame is processed independently at a resolution of
$512\times512$, with depth estimated from the masked frame using
Marigold~\cite{marigold}. Both models are trained with a learning rate of
$10^{-5}$ and a global batch size of 16. MirrorFusion updates only
BrushNet~\cite{brushnet}, whereas MirrorVerse jointly updates BrushNet and the
U-Net~\cite{unet}. During inference, we use 50 denoising steps and a guidance
scale of 7.5.

\subsubsection{VideoPainter}

We initialize VideoPainter~\cite{videopainter} from
CogVideoX-5B-I2V~\cite{cogvideox} and its released context branch, and
fine-tune only the context branch. Training uses clips of up to 49 frames at
$480\times720$, with a learning rate of $10^{-5}$. Following the official protocol, the first frame is generated using FLUX-Fill~\cite{flux} from the masked input without access to the ground-truth
video. During inference, we use 50 denoising steps and a guidance scale of
6.0.

\subsubsection{VACE}

For VACE~\cite{vace}, we fine-tune Wan2.1-VACE-14B using rank-32 LoRA.
Training uses AdamW~\cite{adamw} with a learning rate of $10^{-4}$. During
inference, we use 50 denoising steps, a guidance scale of 5.0, and a VACE
conditioning scale of 1.0.

\section{Metric Definitions}
\label{sec:metric_definitions}

Following the main paper, we evaluate reflection fidelity using PSNR, SSIM, and LPIPS, video quality using FVD and the flow warping error $E_{\mathrm{warp}}$, semantic and shading consistency using DINO similarity and the shading error $\mathrm{SH}_{\mathrm{err}}$, and reflection plausibility using the VLM Judge. The ablation on semantic relational objectives in Table~\ref{tab:relation_placement} of the main paper further reports semantic-category precision, recall, F1, and hallucination rate. We compute all metrics except FVD exclusively within the annotated mirror regions, whereas FVD is evaluated over complete frames without masking.
For PSNR, SSIM, and LPIPS, each method is evaluated at its native output resolution by resizing the ground truth and mask to it via bicubic and nearest-neighbor interpolation, respectively, as in the main paper. For DINO, $\mathrm{SH}_{\mathrm{err}}$, and the semantic-category metrics, we instead resize each generated frame to the ground-truth resolution via bicubic interpolation and sample every fourth frame.
We average these frame-level metrics within each clip and then across all test clips.

\subsection{Peak Signal-to-Noise Ratio (PSNR)}

PSNR measures pixel-level reconstruction fidelity. Following standard practice, we compute the mean squared error exclusively over pixels inside the mirror mask. Higher PSNR indicates more accurate reflection reconstruction.

\subsection{Structural Similarity Index Measure (SSIM)}

SSIM~\cite{ssim} evaluates the structural similarity between generated and ground-truth reflections. We compute the standard full-frame SSIM map and average the values within the mirror mask, excluding regions outside the mirror boundary. Higher SSIM indicates better preservation of structural details.

\subsection{Learned Perceptual Image Patch Similarity (LPIPS)}

LPIPS~\cite{lpips} assesses perceptual discrepancy using deep visual features. We adopt the standard spatial LPIPS implementation, resize the mirror mask to match the spatial feature map resolution, and average the perceptual distances within the masked region. Lower LPIPS indicates higher perceptual similarity to the ground truth.

\subsection{Fr\'echet Video Distance (FVD)}

FVD~\cite{fvd} measures the distributional discrepancy between generated and ground-truth videos. Unlike the masked metrics above, FVD is evaluated over complete frames without mirror masking. We resize these frames to $224\times224$ and compute FVD from their I3D~\cite{I3D} features following the StyleGAN-V~\cite{StyleGAN-V} implementation. Lower FVD indicates superior overall video quality.

\subsection{Flow Warping Error (\texorpdfstring{$E_{\mathrm{warp}}$}{E-warp})}

$E_{\mathrm{warp}}$~\cite{ewarp} evaluates temporal consistency under motion compensation. Specifically, we estimate optical flow between adjacent ground-truth frames using RAFT-Large~\cite{raft} and warp the corresponding generated frames accordingly. We then compute the photometric warping error exclusively over valid mirror-region pixels, discarding out-of-bounds pixels and regions failing forward--backward flow consistency checks. Lower $E_{\mathrm{warp}}$ indicates stronger temporal coherence.

\subsection{DINO Similarity (DINO)}

DINO assesses semantic similarity between generated and ground-truth reflections without requiring strict pixel-level alignment. We resize both frames and extract patch tokens using a frozen DINOv3~\cite{dinov3} ViT-B/16 backbone. Patches with at least 75\% of their area inside the mirror mask are retained as mirror patches. Let $\{\mathbf{p}_{i}\}_{i=1}^{N}$ and $\{\mathbf{g}_{j}\}_{j=1}^{N}$ denote the $\ell_2$-normalized features of the $N$ mirror patches in the generated and ground-truth frames, respectively. We match each patch to its nearest neighbor in the counterpart frame:
\begin{equation}
P=\frac{1}{N}\sum_{i=1}^{N}\max_{j}\,\mathbf{p}_{i}^{\top}\mathbf{g}_{j},\qquad
R=\frac{1}{N}\sum_{j=1}^{N}\max_{i}\,\mathbf{p}_{i}^{\top}\mathbf{g}_{j},
\end{equation}
and compute the harmonic mean $\mathrm{DINO}=2PR/(P+R)$, where $P$ penalizes hallucinated elements absent from the ground truth and $R$ penalizes missing ground-truth elements. Since SRD distills semantic relations from VideoMAEv2~\cite{videomae}, evaluating with DINOv3 avoids feature-space overlap with training supervision. Higher DINO indicates stronger semantic agreement.

\subsection{Shading Error (\texorpdfstring{$\mathrm{SH}_{\mathrm{err}}$}{SH-err})}

$\mathrm{SH}_{\mathrm{err}}$ evaluates whether the generated reflection exhibits low-frequency shading consistent with the ground truth. For each sampled frame, we crop the bounding box of the ground-truth mirror, fill non-mirror pixels with gray, and resize the crop so that its longer side is 640 pixels. Applying the same crop and masking to the generated frame ensures that MoGe-2~\cite{moge2} estimates surface normals of the reflected scene rather than the physical mirror plane. Within the mirror mask eroded by 2 pixels, we fit second-order spherical harmonics (SH) to each crop from its estimated normals via ridge regression:
\begin{equation}
\mathbf{C}=\arg\min_{\mathbf{C}}\,\big\|\mathbf{Y}\mathbf{C}-\mathbf{I}\big\|_{F}^{2}+\lambda M\,\|\mathbf{C}\|_{F}^{2},
\end{equation}
where $\mathbf{Y}\in\mathbb{R}^{M\times 9}$ stacks the nine SH basis functions evaluated at the normals of the $M$ valid mirror pixels, $\mathbf{I}\in\mathbb{R}^{M\times 3}$ denotes their RGB intensities in $[0, 255]$, $\mathbf{C}\in\mathbb{R}^{9\times 3}$ contains the fitted SH coefficients, and $\lambda=10^{-4}$. Each crop yields a reconstructed shading $\hat{\mathbf{s}}(x)=\mathbf{y}(x)^{\top}\mathbf{C}$ at pixel $x$, where $\mathbf{y}(x)$ is the SH basis vector at the normal of $x$, and the error is defined as:
\begin{equation}
\mathrm{SH}_{\mathrm{err}}=\sqrt{\frac{1}{3|\Omega|}\sum_{x\in\Omega}\big\|\hat{\mathbf{s}}^{\mathrm{gen}}(x)-\hat{\mathbf{s}}^{\mathrm{gt}}(x)\big\|_{2}^{2}},
\end{equation}
where $\Omega$ denotes the set of mirror pixels with valid normals in both crops. We compare reconstructed shading rather than raw SH coefficients, as the coefficients become under-constrained when surface normals span only a narrow solid angle. Frames with fewer than 200 valid normals or nearly uniform normals (\eg, a single planar surface) are skipped, and $\mathrm{SH}_{\mathrm{err}}$ is averaged over clips containing at least one valid frame. Lower $\mathrm{SH}_{\mathrm{err}}$ indicates better shading consistency.

\subsection{Semantic-Category Metrics}

The semantic-category metrics in Table~\ref{tab:relation_placement} of the main paper assess whether the generated reflection recovers the ground-truth object and scene categories without introducing unsupported categories. For each sampled frame, we apply OneFormer~\cite{oneformer} with an ADE20K-pretrained Swin-L backbone to three inputs: the ground-truth and generated mirror crops from $\mathrm{SH}_{\mathrm{err}}$, and the full ground-truth frame with the mirror filled in gray. A category is considered present if it covers at least 1\% of the evaluated region and at least 64 pixels, excluding the ADE20K ``mirror'' class. Let $G$, $Q$, and $O$ denote the detected category sets in the ground-truth mirror region, the generated mirror region, and the visible scene, respectively. We compute:
\begin{equation}
\mathrm{Prec.}=\frac{|Q\cap G|}{|Q|},\qquad
\mathrm{Rec.}=\frac{|Q\cap G|}{|G|},
\end{equation}
\begin{equation}
\mathrm{F1}=\frac{2\,\mathrm{Prec.}\cdot\mathrm{Rec.}}{\mathrm{Prec.}+\mathrm{Rec.}},\qquad
\mathrm{Halluc.}=\frac{|Q\setminus(G\cup O)|}{|Q|},
\end{equation}
where $\mathrm{Halluc.}$ measures the fraction of generated categories absent from both the ground-truth reflection and the visible scene. Any metric with an empty denominator is skipped for that frame. Higher precision, recall, and F1, together with a lower hallucination rate, indicate more faithful reflection semantics.

\begin{figure*}[t]
\centering
\begin{tcolorbox}[
    enhanced, width=\linewidth, arc=0pt, frame hidden, colback=white,
    boxsep=0.5mm, left=1mm, right=1mm, top=0.6mm, bottom=1mm,
    interior code={\fill[pattern=crosshatch, pattern color=promptband!30] (interior.south west) rectangle (interior.north east);},
    overlay={\draw[black!80, line width=0.8pt, sketch] (frame.south west) rectangle (frame.north east);}
]
\centering{\ttfamily\bfseries\large VLM Judge Prompt}\par\vspace{-1.6mm}
\begin{tcolorbox}[
    enhanced, arc=0pt, frame hidden, colback=gray!6,
    boxsep=0.5mm, left=1.5mm, right=1.5mm, top=1.5mm, bottom=1.5mm, fontupper=\ttfamily\footnotesize,
    overlay={\draw[black!80, line width=0.6pt, sketch] (frame.south west) rectangle (frame.north east);}
]
\raggedright
\textbf{prompt:} '{}'{}'You are an expert evaluator of video mirror-reflection synthesis. Be strict, objective and consistent. Judge only the region inside the red mirror outline.

\medskip
This is a pure visual judging task. Do not run any commands, do not read or write any files, do not use any tools; answer only from the attached images.

\medskip
The attached images are 2 videos, each given as 16 frames in time order (the same 16 frame indices of a \ph{phblue}{\{N\}}-frame, 15 fps clip in every video; timestamps \ph{phblue}{\{t\_1, ..., t\_16\}}):\\
1. Input (what the model saw; mirror region removed, shown in grey): images 1-16\\
2. Result: images 17-32

\medskip
A video model had to fill in the mirror region (inside the red outline) given the rest of the scene. No ground truth is available. Rate whether the Result is a correct, physically plausible mirror reflection of this scene, on one criterion only.\\
Criterion. \ph{phblue}{\{criterion\}}\\
Score from 0 to 100: 0-20 completely wrong, 21-40 mostly wrong, 41-60 partly correct with clear errors, 61-80 mostly correct with minor errors, 81-100 nearly perfect. If the criterion cannot be judged from the videos, use null.\\
First write one short sentence of reasoning. Answer only with JSON: \{"reason": \ph{phpurple}{"..."}, "score": \ph{phteal}{<integer 0-100 or null>}\}'{}'{}'

\medskip
\#\#\# \ph{phblue}{\{criterion\}} is one of the following (one query each):\\
1. Reflection geometry: are the reflected objects at a plausible position, scale and orientation (including the left-right flip of a mirror) given the surrounding scene and the camera?\\
2. Semantic consistency: does the reflection show objects and a layout that belong to the surrounding scene, without hallucinated or missing content?\\
3. Temporal consistency: does the reflection move consistently with the camera and the scene over time, without flicker, drifting or popping?\\
4. Visual quality: sharpness, realism, absence of blur and artifacts, and seamless blending with the mirror frame.
\end{tcolorbox}
\end{tcolorbox}
\caption{Prompt of the VLM judge. Each criterion is scored in a separate query, with the 32 frames attached as separate images in the order shown. Placeholders in \textcolor{phblue}{\textbf{blue}} are filled in for each clip and criterion, the judge's rationale and score are returned in the \textcolor{phpurple}{\textbf{purple}} and \textcolor{phteal}{\textbf{teal}} fields.}
\label{fig:vlm_prompt}
\vspace{-2mm}
\end{figure*}

\subsection{VLM-as-a-Judge Evaluation}
\label{sec:supp_vlm}

Pixel-level metrics such as PSNR, SSIM, and LPIPS compare each generated result against a single ground-truth reflection and do not directly assess whether the reflected content aligns with the surrounding scene. Here, we detail the blind VLM-as-a-Judge evaluation protocol used for Table~\ref{tab:main_results} of the main paper, where we employ GPT-6.1 Sol without access to ground truth.

\noindent\textbf{Inputs.} We evaluate all five methods from Table~\ref{tab:main_results} of the main paper across the 100 test clips. For each clip and method, the judge receives the masked input, \ie, the ground-truth video with the mirror region filled in grey, alongside the generated video. Both videos consist of 16 uniformly sampled frames extracted at identical temporal indices across all methods. All frames are resized to the ground-truth resolution, cropped to $2\times$ the mirror bounding box, scaled to a longer side of 448 pixels, and overlaid with an identical red mirror outline. To ensure blind evaluation, all frames carry neutral filenames and the prompt omits method identifiers.

\noindent\textbf{Scoring.} Using the prompt in Figure~\ref{fig:vlm_prompt}, the judge scores reflection geometry, semantic consistency, temporal consistency, and visual quality, each queried independently with a fresh context and no tool use. For each query, it outputs a one-sentence rationale followed by an integer score from 0 to 100, or null if the criterion cannot be assessed. We successfully parse all 2,000 responses (100 clips $\times$ 5 methods $\times$ 4 criteria), excluding 7 null cases where the mirror region is too dark or too narrow.

\section{Additional Qualitative Results}
\label{sec:supp_qual}

\subsection{Qualitative Comparisons with Baseline Methods}

\begin{figure*}[t] \centering
    \includegraphics[width=1\textwidth]{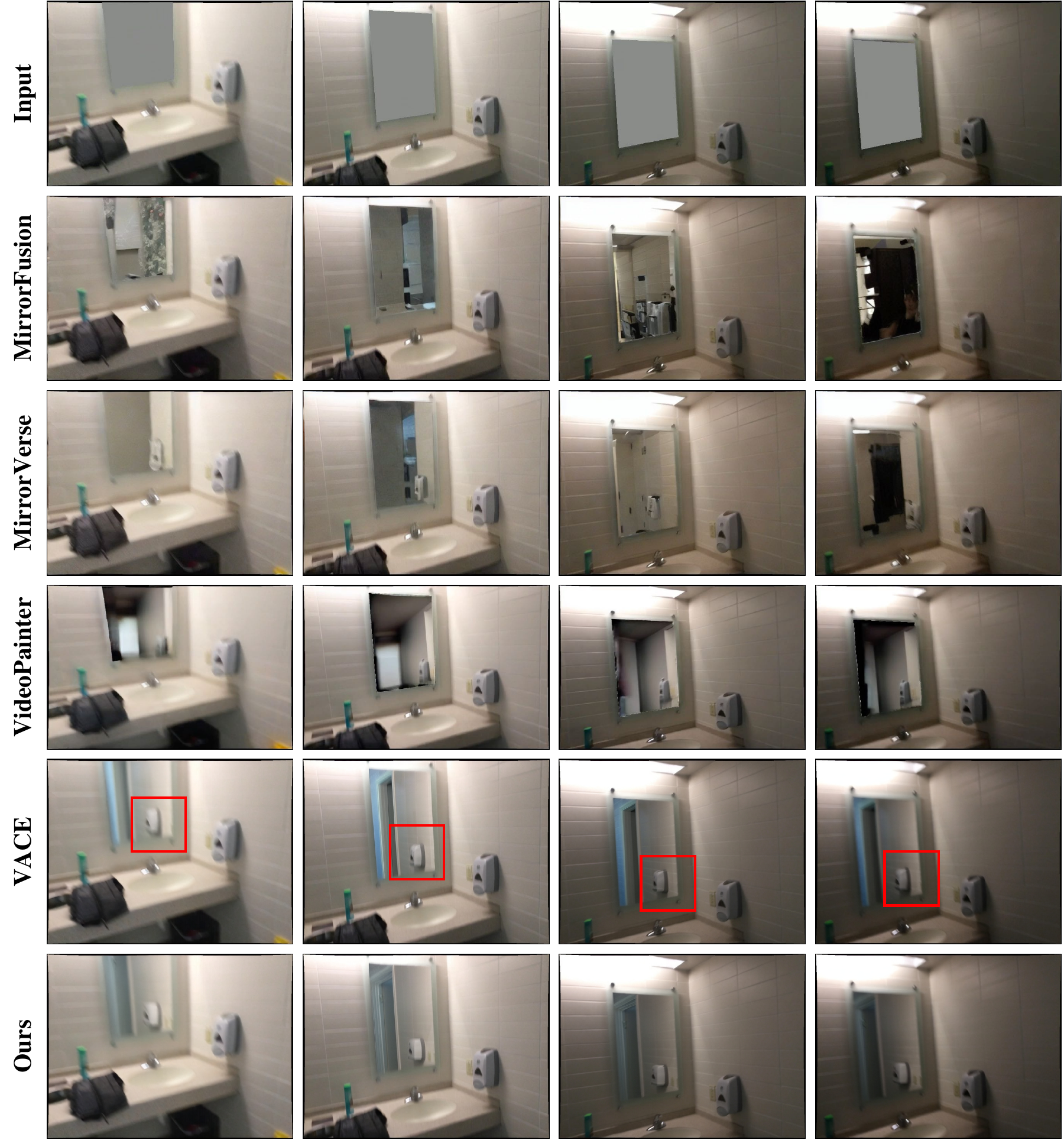}
\caption{\textbf{Additional qualitative comparisons.} Red boxes highlight the missing reflection of the electrical outlet in VACE. MirrorWorld reconstructs the relevant visible elements and maintains their spatial arrangement across frames.}
\label{fig:supp_qualitative_2}
\end{figure*}

\begin{figure*}[t] \centering
    \includegraphics[width=1\textwidth]{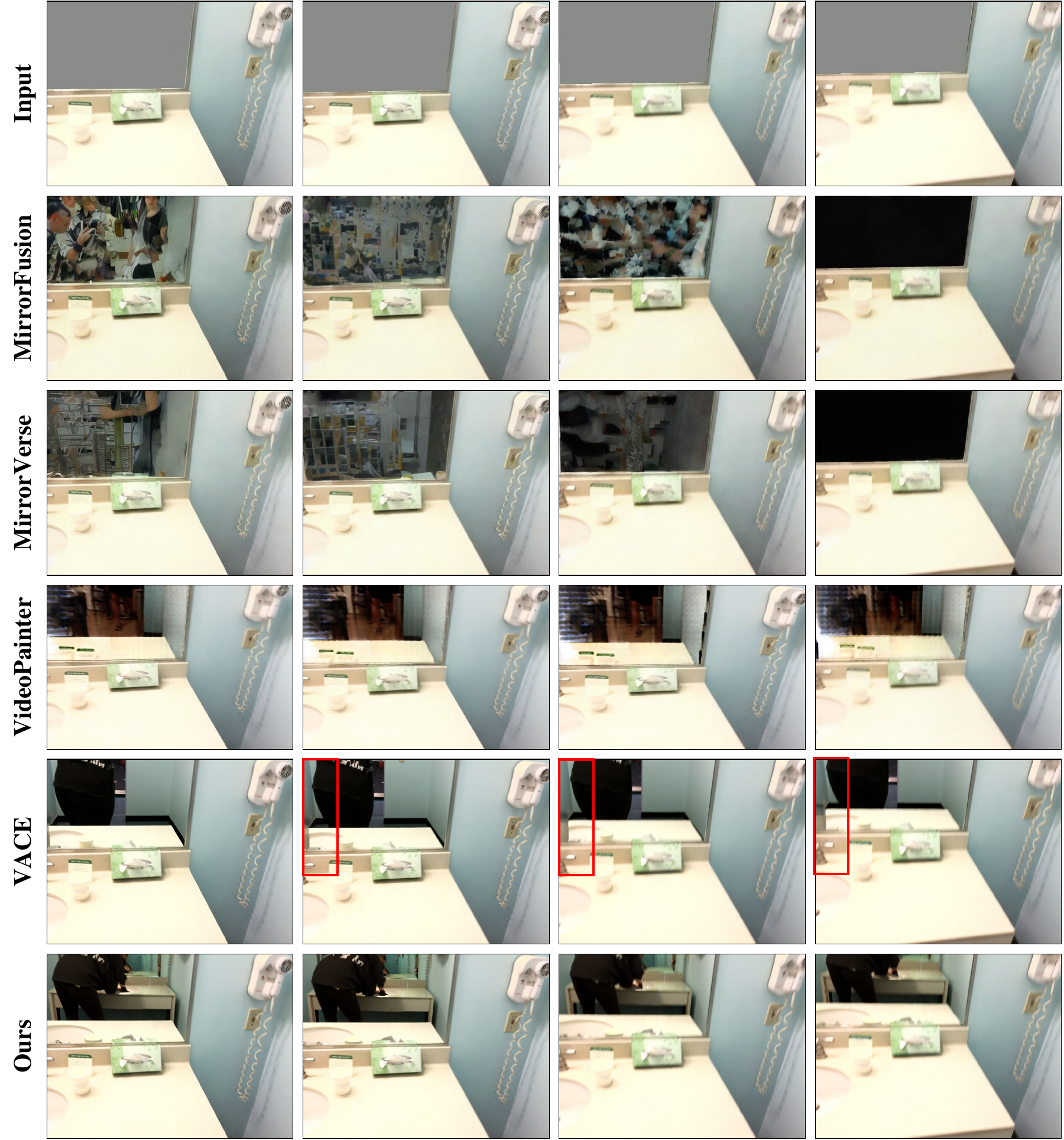}

\caption{\textbf{Additional qualitative comparisons.} Red boxes highlight spatial inconsistencies in VACE. MirrorWorld maintains more coherent reflected content and spatial organization.}

\label{fig:supp_qualitative_1}
\end{figure*}

Figures~\ref{fig:supp_qualitative_2} and \ref{fig:supp_qualitative_1} provide additional comparisons across four sampled frames. MirrorFusion~\cite{mirrorfusion} and MirrorVerse~\cite{mirrorverse} generate unrelated content that changes substantially across frames. VideoPainter~\cite{videopainter} improves temporal coherence, but its generated reflections remain weakly related to the visible scene.

In Figure~\ref{fig:supp_qualitative_2}, the visible wall contains both a dispenser and an electrical outlet. VACE~\cite{vace} reconstructs the dispenser but misses the reflection of the outlet in the regions highlighted by the red boxes. MirrorWorld preserves both elements and maintains their spatial relationship across frames.

In Figure~\ref{fig:supp_qualitative_1}, the image-based methods generate unrelated indoor structures, while VideoPainter produces a temporally coherent but semantically unrelated interior. VACE recovers scene-related content, but the reflected structure near the mirror boundary changes inconsistently across frames. MirrorWorld maintains a more coherent mirror reflection layout. These comparisons are consistent with our design of learning both \emph{what} should be reflected and \emph{how} it should appear within the mirror. SRD associates the mirror with relevant visible content, while GTA guides its reflected arrangement.

\subsection{Qualitative Evaluation}
\label{sec:additional_qualitative}

\begin{figure*}[t] \centering
    \includegraphics[width=1\textwidth]{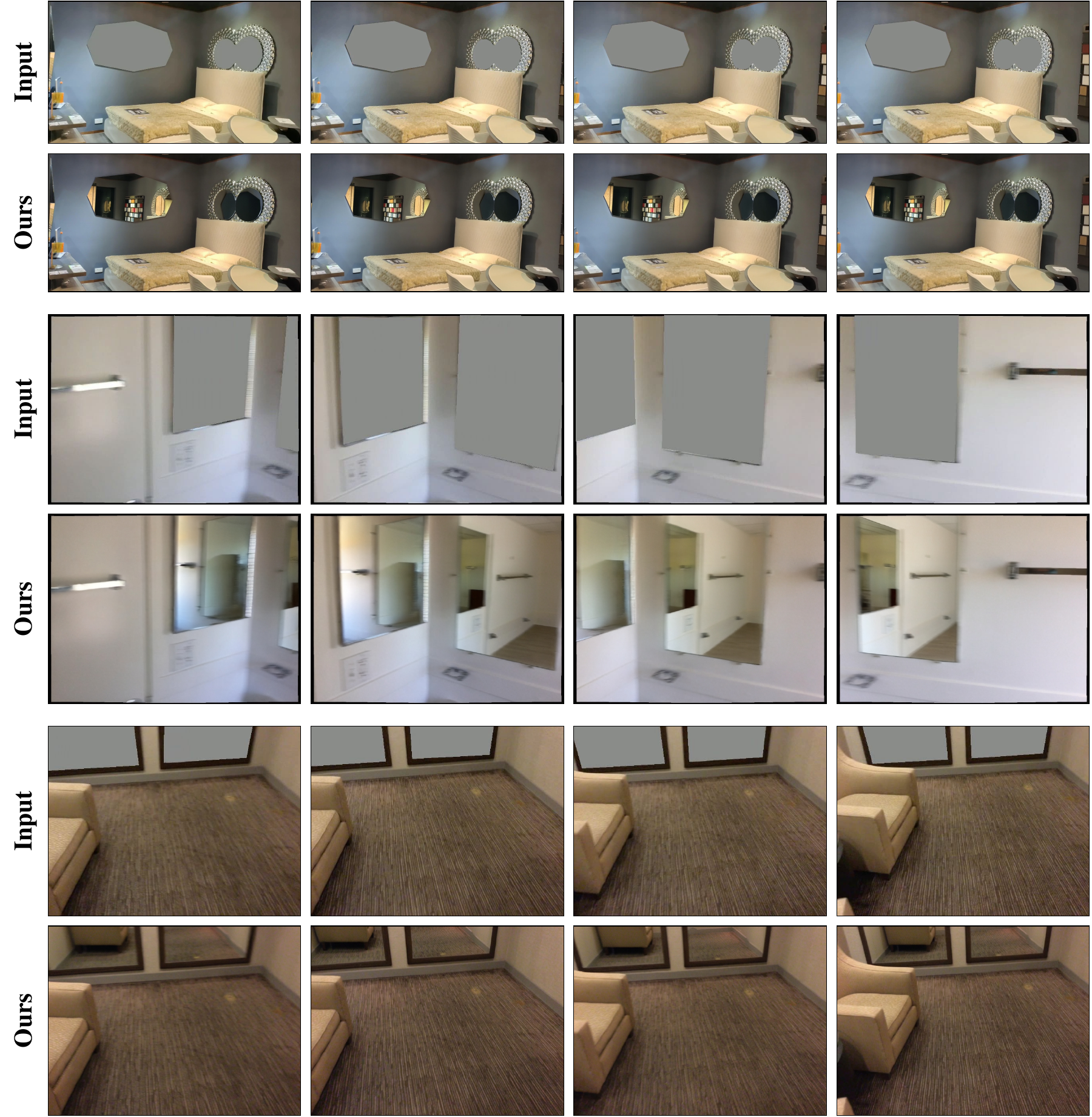}
\caption{\textbf{Qualitative results on scenes with multiple mirrors.} From top to bottom, the examples show a bedroom, a bathroom, and a living room, each with two mirrors. MirrorWorld generates a separate reflection within each mirror and preserves the spatial arrangement of the reflected scene across changing viewpoints.}
\label{fig:supp_multi_mirror}
\end{figure*}

\begin{figure*}[t] \centering
    \includegraphics[width=1\textwidth,height=0.86\textheight,keepaspectratio]{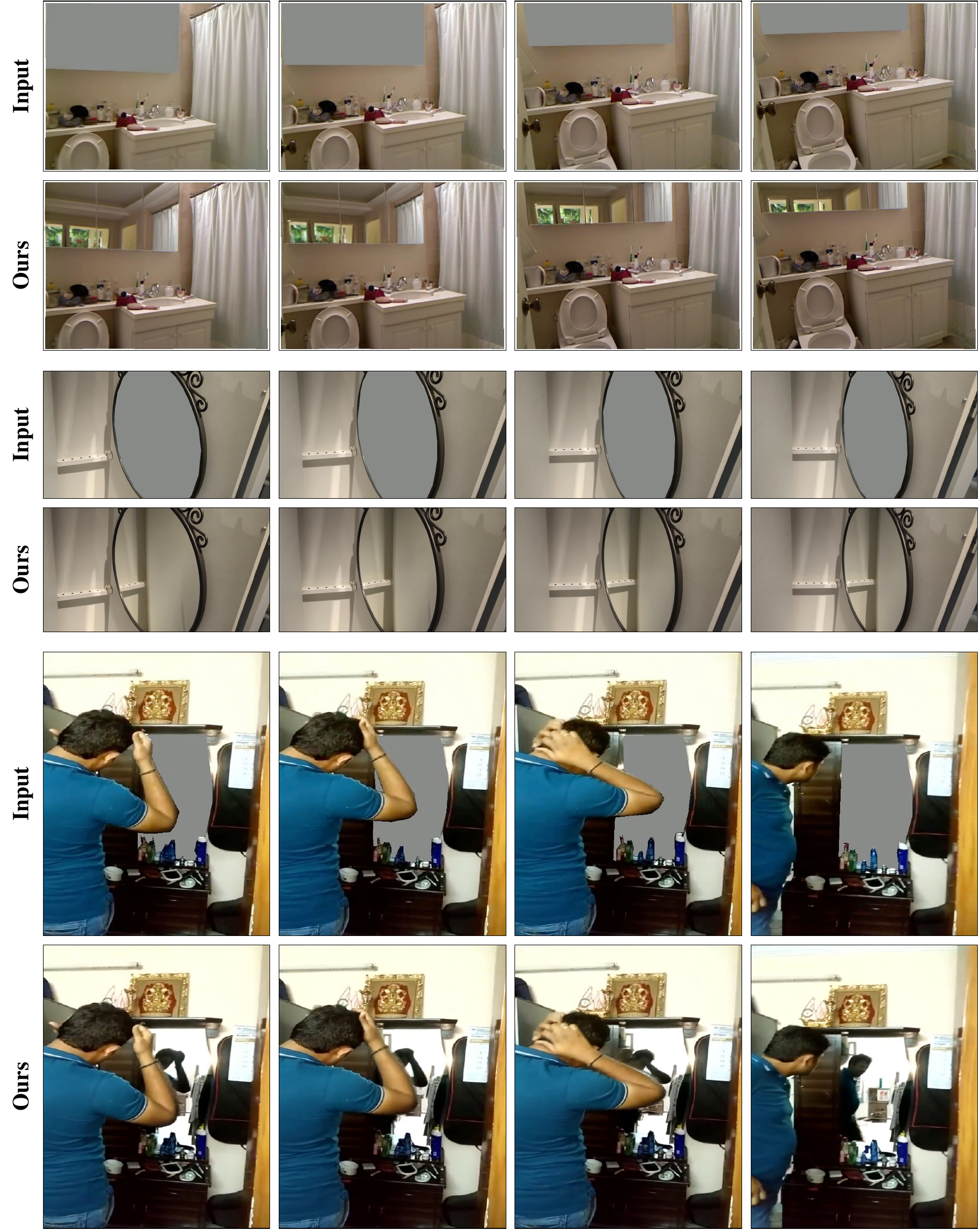}

\caption{\textbf{Additional qualitative results under diverse mirror geometry and motion.} From top to bottom, the examples show an elongated bathroom mirror, an oval mirror, and a mirror with a person moving in front of it. MirrorWorld adapts the reflection to the mirror boundary and maintains the reflected subject as its pose changes across frames.}
\label{fig:supp_geometry_motion}

\end{figure*}

\begin{figure*}[t]
    \centering
    \includegraphics[width=1\textwidth]{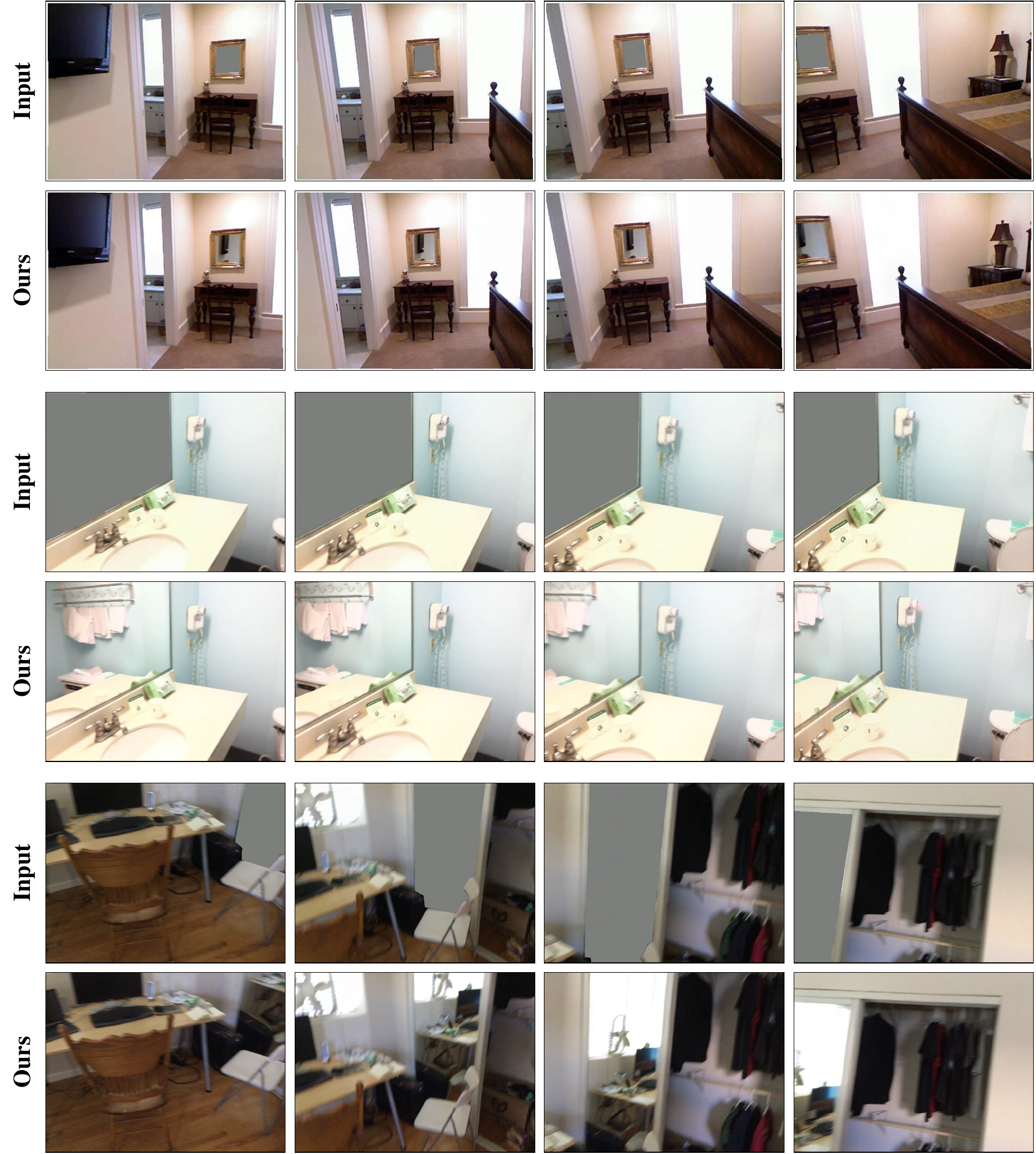}
    \caption{\textbf{Additional qualitative results across diverse indoor scenes.} From top to bottom, the examples show a small bedroom mirror, a larger bathroom mirror, and a mirror in a cluttered room. Each example pairs the masked input with the MirrorWorld result at four matching time points.}
    \label{fig:supp_qualitative_5}
\end{figure*}

Figures~\ref{fig:supp_multi_mirror}--\ref{fig:supp_qualitative_5} present additional qualitative results of MirrorWorld. Each example displays four sampled frames spanning diverse mirror counts, geometries, aspect ratios, scene layouts, and motion patterns.

Figure~\ref{fig:supp_multi_mirror} highlights scenes containing multiple mirrors. In the first example, where a polygonal mirror and an oval mirror are mounted side by side, MirrorWorld synthesizes distinct, viewpoint-consistent reflections within each boundary. In the bathroom scene, reflected doorways and wall boundaries remain temporally stable under camera motion, while the final example preserves the relative spatial layout of the sofa and carpet across frames.

Figure~\ref{fig:supp_geometry_motion} evaluates diverse mirror shapes and dynamic foreground motion. MirrorWorld preserves the reflected window structure within an elongated bathroom mirror throughout the sequence and adapts the synthesized content to an oval mirror boundary. In the final example, the reflection follows variations in the person's pose and position. These results are qualitatively consistent with jointly modeling \emph{what} content to reflect and \emph{how} it is spatially arranged within the mirror.

Figure~\ref{fig:supp_qualitative_5} provides three additional examples across varied mirror scales and scene complexities: a small wall-mounted bedroom mirror, a large bathroom mirror, and a cluttered indoor scene with surrounding furniture and clothing. Across all sampled frames, the synthesized reflections adapt smoothly to changing viewpoints while preserving coherent room geometry, demonstrating the robustness of MirrorWorld across diverse indoor contexts and camera trajectories.

\end{document}